\documentclass[journal]{IEEEtran}
\ifCLASSINFOpdf
\else
\fi

\usepackage{epsfig}
\usepackage{graphicx}
\usepackage{amsmath}
\usepackage{amssymb}
\usepackage{bm}
\usepackage{multirow}
\usepackage{subfigure}
\usepackage{caption}
\usepackage{color}
\usepackage{url}
\usepackage{enumerate}
\usepackage{xspace}
\DeclareMathOperator*{\argmin}{argmin}
\DeclareMathOperator*{\argmax}{argmax}
\makeatletter
\DeclareRobustCommand\onedot{\futurelet\@let@token\@onedot}
\def\@onedot{\ifx\@let@token.\else.\null\fi\xspace}

\def\eg{\emph{e.g}\onedot} 
\def\ie{\emph{i.e}\onedot}

\def\etal{\emph{et al}\onedot}
\makeatother

\begin{document}
%
\title{Learning a Single Tucker Decomposition Network for Lossy Image Compression with Multiple Bits-Per-Pixel Rates}
%
%
%

\author{Jianrui~Cai,
        Zisheng~Cao,
        and~Lei~Zhang,~\IEEEmembership{Fellow,~IEEE}
\thanks{This work is supported by Hong Kong RGC GRF grant (PolyU 152124/15E).}
\thanks{J. Cai and L. Zhang are with Department of Computing, The Hong Kong Polytechnic University, Kowloon Hong Kong (e-mail: $\{$csjcai, cslzhang$\}$@comp.polyu.edu.hk).}
\thanks{Z. Cao is with Camera Group of DJI Innovations Co., Ltd, Shenzhen, China (e-mail: zisheng.cao@dji.com).}}

\maketitle

\begin{abstract}
Lossy image compression (LIC), which aims to utilize inexact approximations to represent an image more compactly, is a classical problem in image processing. 
Recently, deep convolutional neural networks (CNNs) have achieved interesting results in LIC by learning an encoder-quantizer-decoder network from a large amount of data. 
However, existing CNN-based LIC methods usually can only train a network for a specific bits-per-pixel (bpp). 
Such a ``one network per bpp" problem limits the generality and flexibility of CNNs to practical LIC applications. 
In this paper, we propose to learn a single CNN which can perform LIC at multiple bpp rates. 
A simple yet effective Tucker Decomposition Network (TDNet) is developed, where there is a novel tucker decomposition layer (TDL) to decompose a latent image representation into a set of projection matrices and a core tensor. 
By changing the rank of core tensor and its quantization, we can easily adjust the bpp rate of latent image representation within a single CNN. 
Furthermore, an iterative non-uniform quantization scheme is presented to optimize the quantizer, and a coarse-to-fine training strategy is introduced to reconstruct the decompressed images.
Extensive experiments demonstrate the state-of-the-art compression performance of TDNet in terms of both PSNR and MS-SSIM indices.
\end{abstract}
%


\IEEEpeerreviewmaketitle

\section{Introduction}

\IEEEPARstart{A}{s} an indispensable step in many image processing applications, lossy image compression (LIC) is a classical yet still active topic. 
The goal of LIC is to reduce the image storage space without sacrificing much the image quality, and thus provide an economic solution to image storage and transmission systems. 
Recently, with the development of portable imaging devices and social media (\eg, Facebook, Instagram and Flickr), billions of images are transmitted and stored daily on social networks \cite{liu2018}. 
The explosive growth of the amount of shared images on Internet raises higher requirements on LIC for more effective visual communication systems.

A typical LIC system contains mainly three modules: transformation (\eg, an encoder and a corresponding decoder), quantization (e.g., a quantizer), and encoding. 
To compress an image into bitstreams, conventional LIC methods firstly apply predefined transformations to transform an image into a sparse domain, then perform lossy quantization on the transformed coefficients, followed by entropy coding \cite{liu2017random}. 
Notwithstanding their demonstrated success, conventional LIC methods suffer from three major drawbacks. 
First, they generally employ a series of cascaded modules to compress an image, which may introduce cumulative errors because there are few interactions between these modules. 
Second, the transformations employed in these LIC methods are generally designed in a hand-crafted manner (\eg, discrete cosine transform (DCT) for JPEG \cite{shapiro1993embedded} and discrete wavelet transform (DWT) for JPEG 2000 \cite{rabbani2002overview}), which are limited to represent the various complex structures in natural images. 
Third, traditional LIC methods' performance is poor for compression with a low bits-per-pixel (bpp) rate, often generating severe visual artifacts (\eg, blocky artifacts, blurrings and ringings).

Deep convolutional neural networks (CNNs) have recently led to a series of breakthroughs in many vision problems \cite{krizhevsky2012imagenet, girshick2014rich, he2016deep, kim2016accurate, Cai2018deep}.
The flexible non-linear modelling capability and powerful end-to-end training paradigm of CNN also make it a promising new approach to LIC. 
In the last several years, a flurry of CNN-based LIC methods have been proposed, including the study of network structures \cite{toderici2015variable, toderici2017full, balle2017end, theis2017lossy, li2017learning} as well as loss functions \cite{rippel2017real, mentzer2018conditional, johnston2017improved, agustsson2017soft}. 
Firstly, the end-to-end training manner enables CNN-based LIC systems to adaptively learn an effective encoder-decoder pair from a large amount of image data and in a larger context to represent more complex image structures, reducing the artifacts in the decompressed image. 
Secondly, by adopting specific loss functions (\ie, perceptual metrics) in the training, the CNN compressors are able to strengthen certain desired aspects (\ie, perceptual quality) of the decomposed image.

Despite the advantages of employing CNN for compression, there are still some challenges which limit the performance of CNN-based compressors. 
First, existing CNN-based LIC methods can only change the number of latent feature maps and/or quantized values to adjust the bpp rate. 
As a result, the network is trained dedicatedly for a specific bpp rate once at a time. 
Such a ``one network per bpp" problem limits the flexibility and applicability of CNNs to practical image compression systems. 
Second, because of the non-differentiable property of discrete operation, quantizer is hard to be updated during the end-to-end CNN network training. 
Therefore, the optimal decision boundaries of quantization levels are almost unreachable.
Third, existing CNN based LIC methods usually adopt fixed quantization bins to discretize the latent image representation and treat each element of the latent image representation equally. 
Such a quantization scheme, however, ignores the prior knowledge that the local content is spatially variant in an image, and restricts the capability of CNNs in compressing complex image structures.

To address the aforementioned issues, in this work, we propose a new paradigm for deep LIC. 
More specifically, we proposed a deep Tucker Decomposition Network (TDNet) which takes the sparsity/low-rankness of latent image representations into consideration. 
The key component of TDNet is a novel tucker decomposition layer (TDL), which decomposes the latent image representation into a set of projection matrices and a compact core tensor. 
By changing the rank of core tensor and its quantization levels, we can easily adjust the bpp rate of latent image representation, and thus a single CNN model can be trained to compress and reconstruct images under multiple bpp rates. 
Besides, we propose an iterative non-uniform quantization strategy to obtain the optimal quantization boundaries based on the distribution of encoding coefficients. 
A coarse-to-fine training strategy is introduced to train a stable TDNet and reconstruct the decompressed images. 
Extensive experiments demonstrate that, our proposed TDNet trained with the mean-squared error (MSE) loss or the multi-scale structural similarity index (MS-SSIM) \cite{wang2003multiscale} loss can yield competitive results with state-of-the-art CNN-based LIC schemes but it uses only a single network to achieve this goal. 

The contributions of this work are summarized as follows:

\begin{enumerate}[(1)]
\item {We propose an end-to-end trainable deep tucker decomposition network, namely TDNet, which, for the first time to the best of our knowledge, enables a single network to perform LIC at multiple bpp rates.
}

\vspace{1.2mm}
\item {We present an iterative non-uniform quantization scheme to obtain the quantization boundaries of the tensor decomposition coefficients, and adopt a variable-bits quantization scheme to discretize the latent image representation. 
The proposed methods demonstrate state-of-the-art PSNR/SSIM indices and visual quality.}
\end{enumerate}

The remainder of this paper is organized as follows. 
Section \ref{sec:related} provides a brief survey of related work. 
Section \ref{sec:td-net} introduces our proposed TDNet model. 
Section \ref{sec:tdl} presents in detail the tucker decomposition layer. 
Section \ref{sec:all} presents the all-in-one training strategy. 
In Section \ref{sec:expri}, extensive experiments are conducted to evaluate TDNet. 
Finally, several concluding remarks are given in Section \ref{sec:col}.

\section{Related Work}
\label{sec:related}
\subsection{Traditional Lossy Image Compression}
The most prevalent LIC method is JPEG (Joint Photographic Experts Group)\footnote{\url{https://jpeg.org/}}, which first applies discrete cosine transform (DCT) to non-overlapping $8\times8$ image blocks, and then quantizes the transformed DCT coefficients in frequency domain using a predefined quantization table, followed by entropy coding such as Huffman coding and arithmetic coding \cite{wallace1992jpeg}. 
As a significantly improved version of JPEG, JPEG2000 adopts the more powerful discrete wavelet transform (DWT), instead of DCT, to perform time-frequency analysis on images. 
More specifically, JPEG2000 adopts the Cohen-Daubechies-Feauveau (CDF) 9/7 wavelet to decompose an image into multiple bands, and performs scalar-quantization on the DWT coefficients, followed by the Embedded Block Coding with Optimal Truncation (EBCOT) \cite{skodras2001jpeg}.
Another powerful LIC scheme is the so-called Better Portable Graphics (BPG) method\footnote{\url{https://bellard.org/bpg/}}, which is built upon the intra-frame encoding scheme of the High Efficiency Video Coding (HEVC) video compression standard\footnote{\url{https://www.itu.int/rec/T-REC-H.265}}.
It has been proved that BPG can produce smaller files for a given quality than JPEG and JPEG 2000.

Although these traditional LIC approaches have demonstrated their great success, they all adopt hand-crafted transformations to transform the image into some sparse domain for quantization. 
The hand-crafted transformations are limited in adaptively and effectively decomposing complex image structures, resulting in visual artifacts around image edges and textures, especially when the bpp rates are low. 
The deep neural network based LIC methods are then proposed to address these problems.

\subsection{Deep Lossy Image Compression}
Recently, deep neural networks have been investigated and achieved promising results in LIC. 
As a pioneering work, Toderici \etal adopted the recurrent neural network (RNN) to encode and decode images of size 32$\times$32 \cite{toderici2015variable}, and they further extended the network to compress full-resolution images \cite{toderici2017full}. 
Built upon the architecture proposed in \cite{toderici2015variable, toderici2017full}, Johnston \etal \cite{johnston2017improved} modified the recurrent architecture by introducing hidden-state priming to improve spatial diffusion, and replaced the MSE loss by MS-SSIM loss \cite{wang2003multiscale} to increase the visual quality of reconstructed images.

Different from the above methods which employ RNN, methods in \cite{balle2017end, theis2017lossy, li2017learning, rippel2017real, agustsson2017soft, mentzer2018conditional} rely on CNN based auto-encoder architectures. 
Ball{\'e} et al. \cite{balle2017end} used generalized divisive normalization for joint nonlinearity to implement local gain control. 
Li \etal \cite{li2017learning} learned a content-weighted importance map, according to which more bits are allocated to the region with rich content to preserve image edge and texture details. 
Rippel \etal \cite{rippel2017real} aggregated image information across different scales by exploiting the pyramidal decomposition strategy, and introduced the generative adversarial networks (GANs) \cite{goodfellow2014generative} to sharpen the edge of reconstructed images. 
To alleviate the effect of vanishing gradient caused by non-differentiable quantization operation, Theis \etal \cite{theis2017lossy} introduced a smooth approximation of the derivative of the rounding function. 
A soft-to-hard scheme is adopted in \cite{agustsson2017soft} to find assignments to the quantizer.

\begin{figure*}
\begin{center}
\includegraphics[width=1\linewidth]{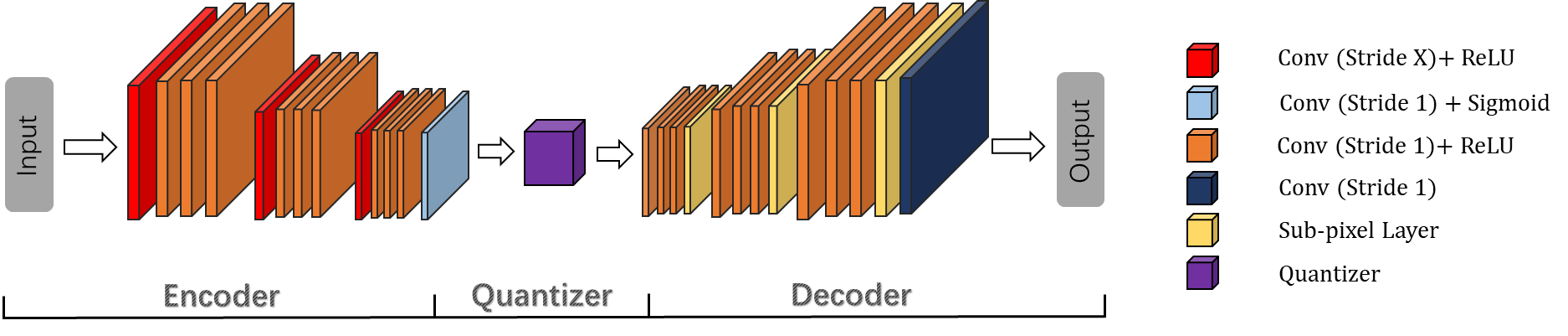}
\end{center}
\caption{Illustration of conventional lossy image compression network architecture.}
\label{fig:Arch1}
\end{figure*}

For all the aforementioned deep LIC methods, the bpp rate of latent image representation can only be adjusted by changing the number of latent feature maps and/or quantized values since the output of encoder should have the same size as the input of decoder. 
Thus, one network can only be trained to deal with a specific bpp rate, making these deep CNN-based LIC methods less flexible. 
In this work, we introduce a novel tucker decomposition layer into CNN, and present a TDNet scheme which enables a single network to tackle with multiple bpp rates for LIC.

\section{Deep Lossy Image Compression Model}
\label{sec:td-net}
In this section, we first summarize the pipeline of conventional CNN-based LIC methods, and then present the pipeline of our proposed TDNet. 
Finally, we present in detail the network architecture.

\subsection{Overview of Conventional LIC Network Pipeline}
Existing deep LIC networks can be generally formulated as a joint rate-distortion optimization process to learn an encoder, a quantizer, and a decoder. 
The architecture of those networks is shown in Figure \ref{fig:Arch1}. 
Given a set of training images $\{\bm{x}_i\}{_{i=1}^N}$, where $N$ is the total number of training images, deep LIC methods aim to learn a nonlinear analysis transformation encoder $E(\cdot)$, a quantizer $Q(\cdot)$, and a nonlinear synthesis transformation decoder $D(\cdot)$. 
The encoder $E(\cdot)$ first converts an input image $\bm{x}_i$ into a latent feature representation $\bm{z}_i=E(\bm{x}_i)$. 
Then, the quantizer $Q(\cdot)$ quantizes the features into discrete values $\hat{\bm{z}}_i=Q(\bm{z}_i)$, which can be losslessly encoded into a bitstream for transmission or storage. 
Once the bitstream is received by the decoder $D(\cdot)$, an approximation of the original image is obtained as $\hat{\bm{x}}_i=D(\hat{{\bm{z}}_i})$. 
Overall, the deep image compression pipeline can be formulated as:
\begin{equation}
\begin{aligned}
\hat{\bm{x}_i} = D ( Q ( E ( \bm{x}_i, \Omega ) ), \Theta ),
\end{aligned}
\end{equation}
where $\Omega$ and $\Theta$ are the parameters of encoder $E(\cdot)$ and decoder $Q(\cdot)$, respectively. 

Given a certain compression ratio, the network is expected to learn the parameters $\Omega$ and $\Theta$ to minimize the distortion of the reconstructed image. 
Note that the compression ratio can be defined as $\alpha=\frac{C(\bm{x}_i)}{C(\bm{z}_i)}$, where $C(\cdot)$ is the function to calculate the average number of bits to store a pixel of an image.
Since $C(\bm{x}_i)$ is usually a constant for the original image $\bm{x}_i$ without compression, we can adjust $C(\bm{z}_i)$ to change the compression ratio $\alpha$. 
For most of the existing CNN based LIC methods, one can only change the number of feature maps and quantization levels to adjust $C(\bm{z}_i)$ of latent image representation $\bm{z}_i$. 
As a result, usually a specific network has to be trained for a certain compression ratio, or bpp rate. 
For a new bpp rate, a new network has to be trained by adjusting the number of latent representation feature maps and quantization levels.

\subsection{Proposed LIC Network Pipeline}
%
\begin{figure*}
\begin{center}
\includegraphics[width=1\linewidth]{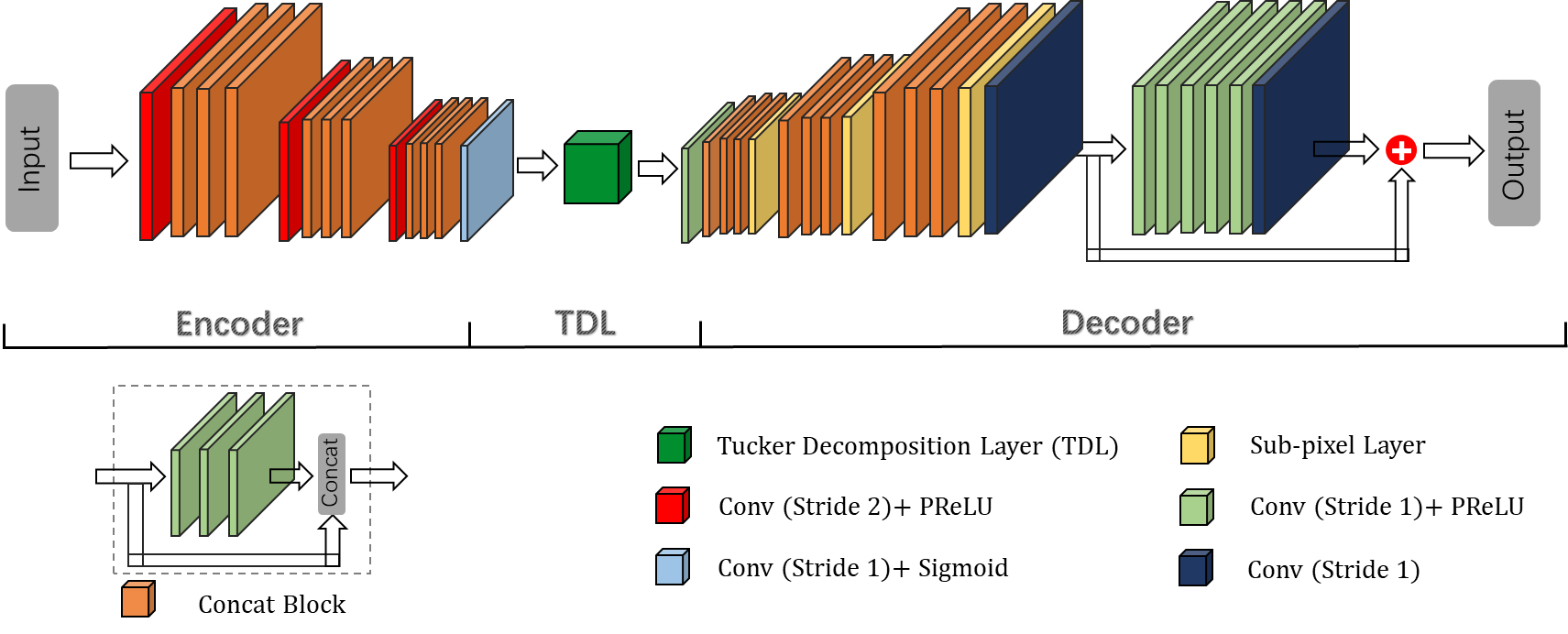}
\end{center}
\caption{Illustration of our proposed TDNet architecture. }
\label{fig:Arch}
\end{figure*}

Our proposed TDNet is designed to achieve the objective of multiple bpp rates with a single network. 
The pipeline of TDNet is shown in Figure \ref{fig:Arch}. 
Instead of directly quantizing the latent image representation into a bitstream as in conventional deep LIC methods, we introduce a novel tucker decomposition layer (TDL) to process the latent image representation.
Denote by $T(\cdot)$ the decomposition operation of TDL, and by $T^{-1}(\cdot)$ the inverse operation.
Given the latent image representation $\bm{z}_i$, we use $\{\bm{Y}, \bm{U}^{(1)}, \bm{U}^{(2)}, \bm{U}^{(3)}\}=T(\bm{z}_i)$ to decompose the features into 3 orthogonal matrices $\{\bm{U}^{(n)}\}{_{n=1}^3}$ and a core tensor $\bm{Y}$, and then quantize the decomposed components to generate bitstream. 
Once the bitstream is received, with $T^{-1}(\cdot)$ we can de-quantize and reproduce the features by back-projecting the core tensor and 3 orthogonal matrices into the approximation $\hat{\bm{z}}_i = T^{-1}(\bm{Y}, \bm{U}^{(1)}, \bm{U}^{(2)}, \bm{U}^{(3)})$.

By changing the rank of core tensor in TDL, we can easily adjust the bpp rates and hence the compression ratio $\alpha$ while keeping the size of latent image representation unchanged. 
Once $\hat{\bm{z}}_i$ is received, we can obtain an approximation of the original input image by $\hat{\bm{x}_i}=R(D(\hat{\bm{z}}_i))$, where $R(\cdot)$ is the reconstruction network to reproduce the decompressed image, and $D(\cdot)$ is the deconvolutional process to up-sample the latent image representations to the size of original images. 
Together, the decoder can be presented as $R(D(\cdot))$. The pipeline of the proposed TDNet can be formulated as:
\begin{equation}
\begin{aligned}
\hat{\bm{x}_i} = R(D(T^{-1}(T(E(\bm{x}_i, \Omega ))), \Theta ), \Pi),
\end{aligned}
\end{equation}
where $\Omega$ is the parameter of encoder $E(\cdot)$, and $\Theta$ and $\Pi$ are the parameters of decoder $D(\cdot)$ and $R(\cdot)$, respectively. 
Being optimized in an end-to-end manner, the network is expected to learn the parameters $\{\Omega, \Theta, \Pi\}$ to minimize the distortion of the reconstructed image.

\subsection{Architecture of TDNet}
\noindent \textbf{Loss Function:}
The loss function of a LIC network defines how close or how similar the decompressed image is to the original image. 
Many existing deep CNN based LIC methods \cite{mentzer2018conditional, agustsson2017soft} use the perceptual loss such as the MS-SSIM loss \cite{wang2003multiscale, zhao2017loss} to strength the perceptual quality of the compressed image. 
The MS-SSIM loss can also be adopted into our TDNet to learn the LIC network. 
Refer to Figure \ref{fig:Arch}, we require both the deconvolution output and the final output of the network to be similar to the original image, resulting in the following loss function:
\begin{equation}\label{msssim}
\begin{aligned}
l_\text{MS-SSIM}(\Omega,& \Theta, \Pi)=1-\text{MS-SSIM}(D(T^{-1}(T(E(\bm{x}_{i}, \Omega))), \Theta))\\
&-\lambda\cdot\text{MS-SSIM}(R(D(T^{-1}(T(E(\bm{x}_{i}, \Omega))), \Theta), \Pi)),
\end{aligned}
\end{equation}
where $\bm{x}_{i}$ refers to the $i$-th image in $\{\bm{x}_i\}_{i=1}^N$, and $\lambda$ is a parameter to balance the loss between the intermediate deconvolution output and the final reconstruction.

Considering that most of the classical LIC methods such as JPEG and JPEG2000 take MSE as the objective to optimize, it is also important to validate whether a deep LIC network can achieve good MSE or equivalently PSNR measures. 
By minimizing the MSE of both the deconvolution output and the final output of the network, the MSE oriented loss function of the proposed network can be formulated as:
\begin{equation}\label{l2}
\begin{aligned}
l_\text{MSE}(\Omega,& \Theta, \Pi) 
=\frac{1}{N}\sum_i^N\|\bm{x}_{i} - D(T^{-1}(T(E(\bm{x}_{i}, \Omega))), \Theta)\|_{2}\\
&+\frac{\lambda}{N}\sum_i^N\|\bm{x}_{i} - R(D(T^{-1}(T(E(\bm{x}_{i}, \Omega))), \Theta), \Pi)\|_{2}.
\end{aligned}
\end{equation}

\vspace{1.2mm}
\noindent \textbf{Encoder:} 
Our encoder network consists of 3 types of layers, which are shown with 3 different colors in Figure \ref{fig:Arch}. 
Instead of using Rectified Linear Unit (ReLU), we adopt Parametric Rectified Linear Units (PReLU) \cite{he2015delving} as the activation function since it could improve the model fitting capability with little extra computational cost. 
Several convolution layers with a stride of 2 are utilized to downsample the feature maps, and the sigmoid function is used to project the data into the range of $[0, 1]$. 
Besides, Concat operations are adopted to concatenate the feature maps of two layers to ensure maximum information flow. 
By stacking several convolutional layers, PReLU and Concat layers, the encoder network can transform the images into a compact domain with reduced redundancy. 
The detailed settings of the encoder network are summarized in Table \ref{tab1}.

\begin{table}
\normalsize
\caption{Encoder network architecture.}
\begin{center}
\begin{tabular}{c||c}
  \hline \hline
  \textbf{Layer}                                   &\textbf{Activation size}                      \\
  \hline \hline   
  Input                                            &  $320\times320\times3$                       \\
  \hline
 \multirow{2}{*}{}
  Conv + PReLU                                     &  \multirow{2}{*}{$160\times160\times64$ }    \\  
  ($3\times3\times64$, stride 2, pad 1)                                                           \\
  \hline
  \multirow{2}{*}{}
  Concat blocks $\times 3$                         &  \multirow{2}{*}{$160\times160\times256$}    \\
  ($3\times3\times64$, stride 1, pad 1)                                                           \\
  \hline
 \multirow{2}{*}{}
  Conv + PReLU                                     &  \multirow{2}{*}{$80\times80\times128$}      \\
  ($3\times3\times128$, stride 2, pad 1)                                                          \\
  \hline
  \multirow{2}{*}{}
  Concat blocks $\times 3$                         &  \multirow{2}{*}{$80\times80\times320$}      \\
  ($3\times3\times64$, stride 1, pad 1)                                                           \\
  \hline
  Conv + PReLU                                     &  \multirow{2}{*}{$40\times40\times256$}      \\
  ($3\times3\times256$, stride 2, pad 1)                                                           \\
  \hline
  \multirow{2}{*}{}
  Concat blocks $\times 3$                         &  \multirow{2}{*}{$40\times40\times640$}      \\
  ($3\times3\times128$, stride 1, pad 1)                                                           \\
  \hline
  \multirow{2}{*}{}
  Conv + Sigmoid                                   &  \multirow{2}{*}{$40\times40\times32$}    \\
  ($3\times3\times32$, stride 1, pad 1)                                                    \\
  \hline
\end{tabular}
\end{center}
\label{tab1}
\end{table}
\vspace{1.2mm}
\noindent \textbf{Tucker Decomposition Layer (TDL):}
Our TDL consists of two operations: $T(\cdot)$, which decomposes and quantizes the features into 3 orthogonal matrices $\{\bm{U}^{(n)}\}_{n=1}^3$ and a core tensor $\bm{Y}$, and $T^{-1}(\cdot)$, which de-quantizes and projects the core tensor and 3 orthogonal matrices back into the features. 
By setting the ranks $\{R_1, R_2, R_3\}$ of the three matrices and setting the quantization level $M$ of the core tensor, the compression ratio of the network can be calculated as:
\begin{equation}\label{rate}
\begin{aligned}
\alpha = \frac{C(\bm{x}_i)}{C(\bm{Y})+C(\bm{U}^{(1)})+C(\bm{U}^{(2)})+C(\bm{U}^{(3)})}.
\end{aligned}
\end{equation}

To change the compression rate, we can adjust the ranks of the decomposition matrices and the quantization levels $\{R_1, R_2, R_3, M\}$ of the core tensor, instead of retraining the network, and consequently achieve the goal of multiple bpp rates with a single network. 
More detail of the proposed TDL can be found in Section \ref{sec:tdl}.

To ensure the end-to-end training of the network, the gradient of each component should be calculated for back-propagation. 
Though the tucker decomposition operation is non-differentiable and we cannot differentiate it with respect to its argument, based on the straight through estimator on gradient in \cite{theis2017lossy}, fortunately, we could set the derivative of tucker decomposition layer as:
\begin{equation}\label{dtkl}
\left\{
\begin{aligned}
&\frac{d}{d\bm{z}}T(\bm{z}) = 1;\\
&\frac{\partial}{\partial\bm{Y}}T^{-1}(\bm{Y}, \bm{U}^{(n)}) = 1;\\
&\frac{\partial}{\partial\bm{U}^{(n)}}T^{-1}(\bm{Y}, \bm{U}^{(n)}) = 1. 
\end{aligned}
\right.
\end{equation}
By setting the derivative to 1, the network can back propagate the loss from a decoder to an encoder. 
Thus, the whole network can be trained in an end-to-end manner.

\begin{table}
\normalsize
\caption{Decoder network architecture.}
\begin{center}
\begin{tabular}{c||c}
  \hline \hline
  \textbf{Layer}                                   &\textbf{Activation size}                       \\
  \hline \hline   
  Input                                            &  $40\times40\times32$                         \\
  \hline
 \multirow{2}{*}{}
  Conv + PReLU                                     &  \multirow{2}{*}{$40\times40\times256$ }      \\  
  ($3\times3\times256$, stride 1, pad 1)                                                           \\
  \hline
 \multirow{2}{*}{}
  Concat blocks $\times 3$                         &  \multirow{2}{*}{$40\times40\times640$}       \\
  ($3\times3\times128$, stride 1, pad 1)                                                           \\
  \hline
 \multirow{2}{*}{}
  Sub-pixel                                        &  \multirow{2}{*}{$80\times80\times160$}      \\
  (upsampling factor: $\times2$)                                                                  \\
  \hline
  \multirow{2}{*}{}
  Concat blocks $\times 3$                         &  \multirow{2}{*}{$80\times80\times352$}      \\
  ($3\times3\times64$, stride 1, pad 1)                                                           \\
  \hline
  Sub-pixel                                        &  \multirow{2}{*}{$160\times160\times88$}      \\
  (upsampling factor: $\times2$)                                                                   \\
  \hline
  \multirow{2}{*}{}
  Concat blocks $\times 3$                         &  \multirow{2}{*}{$160\times160\times280$}      \\
  ($3\times3\times64$, stride 1, pad 1)                                                             \\
  \hline
  \multirow{2}{*}{}
   Sub-pixel                                       &  \multirow{2}{*}{$320\times320\times70$}       \\
  (upsampling factor: $\times2$)                                                                   \\
  \hline
  \multirow{2}{*}{}
  Conv                                             &  \multirow{2}{*}{$320\times320\times3$}    \\  
  ($3\times3\times3$, stride 1, pad 1)                                                           \\
  \hline
  \multirow{2}{*}{}
  Conv + PReLU $\times5$                           &  \multirow{2}{*}{$320\times320\times64$}    \\  
  ($3\times64\times3$, stride 1, pad 1)                                                           \\
  \hline
  \multirow{2}{*}{}
  Conv                                             &  \multirow{2}{*}{$320\times320\times3$}    \\  
  ($3\times3\times3$, stride 1, pad 1)                                                           \\
  \hline
  Residual Sum                                     &  $320\times320\times3$                         \\
  \hline
\end{tabular}
\end{center}
\label{tab2}
\end{table}
\vspace{1.2mm}
\noindent \textbf{Decoder:}
Our decoder network consists of a deconvolutional sub-network and a reconstruction sub-network, as shown in Figure \ref{fig:Arch}. 
The deconvolutional sub-network basically mirrors the architecture of the encoder, and the stride of all convolutional layers is set to $1$ since there is no need to downsample the feature maps. 
To ensure that the output image will have the same size as the input one, the sub-pixel layer \cite{shi2016real} is adopted to reshape and upsample feature maps. 
Usually, the deconvolution sub-network can deliver a rough approximation of the original image. 
The reconstruction sub-network aims to further enhance the deconvlution output by reproducing the missing details and textures in the encoding and TDL quantization process. 
Inspired by \cite{kim2016accurate, zhang2017beyond}, we propose to use the residual learning framework for the design of reconstruction sub-network. 
It consists of five convolutional layers, PReLUs and an element-wise addition operation. 
With the reconstruction sub-network, the final output image quality can be much refined. 
The detailed settings of the decoder network are summarized in Table \ref{tab2}.

\begin{figure*}[t]
\begin{center}
\includegraphics[width=1\linewidth]{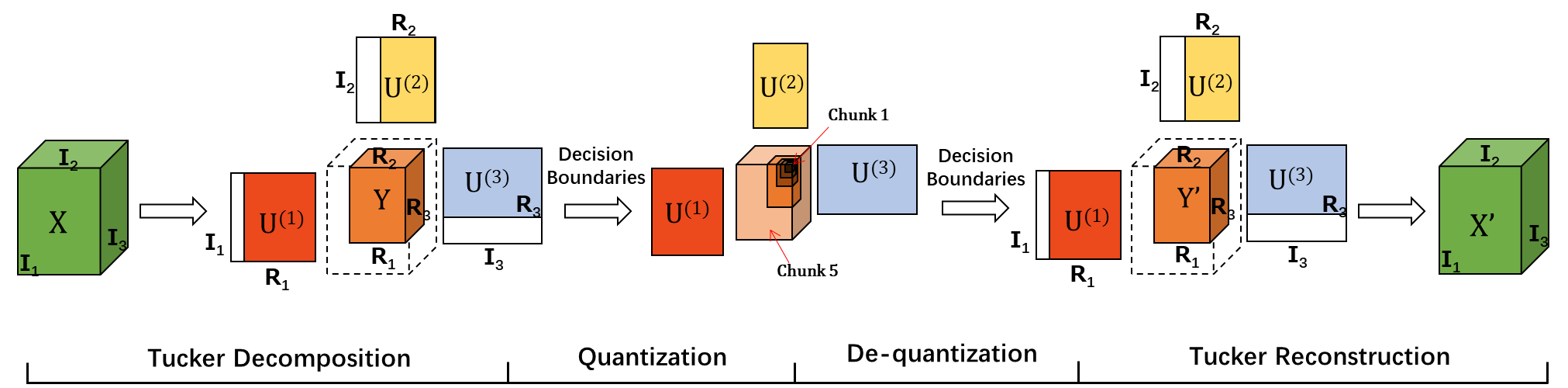}
\end{center}
\caption{Flowchart of the proposed tucker decomposition layer.}
\label{fig:tucker}
\end{figure*}

\section{Tucker Decomposition Layer}
\label{sec:tdl}
In this section, we first introduce some necessary notations and preliminaries of tensor decomposition, and then present in detail the proposed TDL.

\subsection{Notations and Preliminaries}
Denote by $\bm{A} \in \mathbb{R}^{I_1 \times I_2 \times \cdots \times I_N}$ an $N$-order tensor, and denote by $a_{i_1 \cdots i_n \cdots i_N}$ its elements, where $1 \leq i_n \leq i_N$. Let $\bm{B} \in \mathbb{R}^{J_n \times I_n}$ denote a matrix. 
The \emph{mode-$n$ product} of a tensor $\bm{A}$ and a matrix $\bm{B}$ can be defined as \cite{kolda2009tensor}:
\begin{equation}\label{moden1}
\begin{aligned}
\bm{C} = \bm{A} \times_{n} \bm{B} ,
\end{aligned}
\end{equation}
where the symbol $\times_{n}$ denotes the tensor-times-matrix operation, and the \emph{mode-$n$ product} output $\bm{C} \in \mathbb{R}^{I_1 \times \cdots \times I_{n-1} \times J_n \times I_{n+1} \times \cdots \times I_N}$ is a tensor of order $N$. 
The elementwise representation of Eq. (\ref{moden1}) can be written as:
\begin{equation}\label{moden2}
\begin{aligned}
&\bm{C}(i_1,\cdots, i_{n-1}, j_n, i_{n+1}, \cdots, i_N) =\\
&\sum_{k=1}^{I_n}\bm{A}(i_1,\cdots, i_{n-1}, k, i_{n+1}, \cdots, i_N)~\bm{B}(j_n, k).
\end{aligned}
\end{equation}

The \emph{mode-$n$ product} can also be calculated by matrix multiplication:
\begin{equation}\label{moden3}
\begin{aligned}
\bm{C}_{(n)} = \bm{B}\bm{A}_{(n)} ,
\end{aligned}
\end{equation}
where $\bm{A}_{(n)} = unfold_n(\bm{A}) \in \mathbb{R}^{I_n \times (I_1 \cdots I_{n-1} I_{n+1}\cdots I_N)}$ and $\bm{C}_{(n)} = unfold_n(\bm{C}) \in \mathbb{R}^{J_n \times (I_1 \cdots I_{n-1} I_{n+1}\cdots I_N)}$, $1 \leq n \leq N$, are called \emph{mode-$n$ matrices}. 
Note that the operator $unfold_n(\cdot)$ is the process of reordering the elements of an $n$-way date array into a matrix. 
Conversely, the unfolding matrices along the $n^{th}$ mode can be transformed back to the tensor by the $unfold_n(\cdot)$ operation.

For convenience, we define $\bm{A} \bar{\times}_{-n}\{\bm{B}^{(j)}\}{_{j=1}^N}$ as \cite{kolda2009tensor}:
\begin{equation}\label{moden4}
\begin{aligned}
&\bm{A} \bar{\times}_{-n}\{\bm{B}^{(j)}\}{_{j=1}^N}\\
&=\bm{A} \times_1 \bm{B}^{(1)} \times_2 \cdots \times_{n-1} \bm{B}^{(n-1)} \times_{n+1} \bm{B}^{(n+1)} \cdots \times_N \bm{B}^{(N)}\\
&= \bm{A}_{(n)}(\bm{B}^{(N)} \otimes \cdots \otimes \bm{B}^{(n+1)} \otimes \bm{B}^{(n-1)} \otimes \cdots \otimes \bm{B}^{(1)}),
\end{aligned}
\end{equation}
where $\otimes$ denotes the \emph{Kronecker product}. 
The SVD of $\bm{A}_{(n)}$ is defined as:
\begin{equation}\label{moden5}
\begin{aligned}
\bm{A}_{(n)} = \bm{\Psi}^{(n)}\bm{\Sigma}^{(n)}{\bm{V}^{(n)}}^{T},
\end{aligned}
\end{equation}
and the leading $R_n$-dimensional left singular subspace of $\bm{A}_{(n)}$ is defined as $\bm{\Psi}^{(n)}_{r_n} = \bm{\Psi}^{(n)}(:,1:R_n)$.

\subsection{Tucker Decomposition Layer}
As a powerful low rank approximation approach, tensor decomposition, \eg, Tucker decomposition \cite{tucker1966some} and CP decomposition \cite{hitchcock1927expression}, has been successfully used in various tasks, such as multispectral image restoration \cite{xie2016multispectral, xie2017kronecker}, 3D image reconstruction \cite{sauve19993d}, and higher-order web link analysis \cite{kolda2005higher}.
Inspired by the success of tensor decomposition methods, we introduce a novel TDL into the network architecture to achieve the goal that a single LIC network can perform image compression with multiple bpp rates. 
The flowchart of the proposed TDL is illustrated in Figure \ref{fig:tucker}. 
It consists of 4 major components: 1) tucker decomposition; 2) quantization; 3) de-quantization; and 4) tucker reconstruction. 
The details of each component are described as follows.

\vspace{1.2mm}
\noindent \textbf{Tucker decomposition:} 
Tucker decomposition aims to decompose an $N$-order tensor $\bm{X} \in \mathbb{R}^{I_1 \times I_2 \times \cdots \times I_N}$ as an affiliation of $N$ orthogonal bases $\{\bm{U}^{(n)} \in \mathbb{R}^{I_n \times R_n}\}{_{n=1}^N}$ and the associated core tensor $\bm{Y} \in \mathbb{R}^{R_1 \times R_2 \times \cdots \times R_N}$, where $R_n \leq I_n$. 
It can be formulated as:
\begin{equation}\label{tucker1}
\begin{aligned}
&\bm{Y} = \bm{X} \times_1 \bm{U}^{(1)^T} \times_2 \bm{U}^{(2)^T} \times_3 \cdots \times_N \bm{U}^{(N)^T}\\
\Leftrightarrow~~ &\bm{X} \approx \hat{\bm{X}} = \bm{Y} \times_1 \bm{U}^{(1)} \times_2 \bm{U}^{(2)} \times_3  \cdots \times_N \bm{U}^{(N)}.
\end{aligned}
\end{equation}

To find the optimal orthogonal matrices $\{\bm{U}^{(n)}\}{_{n=1}^N}$ and the core tensor  $\bm{Y}$, we could minimize the error between the original data tensor $\bm{X}$ and its approximation $\hat{\bm{X}}$, leading to the following optimization problem:
\begin{equation}\label{tucker2}
\begin{aligned}
{\mathop{\mathop{\argmin}_{{\textbf{U}^{(1)}}, \textbf{U}^{(2)}, \cdots, \textbf{U}^{(N)}}}_ \textbf{Y}}
\|\bm{X} - \bm{Y} \times_1 \bm{U}^{(1)} \times_2 \bm{U}^{(2)} \times_3 \cdots \times_N \bm{U}^{(N)}\|_F^2.
\end{aligned}
\end{equation}
Since $\|\bm{X}\|_F^2$ is a constant, according to \cite{de2000multilinear, kolda2009tensor, andersson1998improving, kolda2006multilinear}, Eq. (\ref{tucker2}) can be recast as an optimization problem to maximize $\|\bm{Y}\|_F^2$. We have:
\begin{equation}\label{tucker3}
\begin{aligned}
{\mathop{\argmax}_{{\textbf{U}^{(1)}}, \textbf{U}^{(2)}, \cdots, \textbf{U}^{(N)}}}
\|\bm{X} \times_1 \bm{U}^{(1)^T} \times_2 \bm{U}^{(2)^T} \times_3 \cdots \times_N \bm{U}^{(N)^T}\|_F^2.
\end{aligned}
\end{equation}

To solve Eq. (\ref{tucker3}), we first employ the higher order singular value decomposition (HOSVD) \cite{de2000multilinear} to initialize a set of basis factor matrices $\{\bm{U}^{(n)}_{0}\}{_{n=1}^N}$, then utilize the higher order orthogonal iteration (HOOI) \cite{de2000best} to iteratively update the orthogonal matrices ${\{{\{\bm{U}^{(n)}_{k}\}}{_{n=1}^N}\}}{^S_{s=1}}$ until convergence, where $s$ is the index of loop. 
With the obtained set of optimal orthogonal matrices $\{\bm{U}^{(n)}\}{_{n=1}^N}$, we can easily obtain the corresponding core tensor $\bm{Y}$ by Eq. (\ref{tucker1}).

Specifically, for our TDNet the order of the feature tensor is $N=3$. 
Given a $3$-order tensor $\bm{X} \in \mathbb{R}^{I_1 \times I_2 \times \times I_3}$ and the desired rank of output $\{R_1, R_2, R_3\}$, we first compute the leading $R_n$-dimensional left singular subspace of $\bm{X}_{(n)}$ to initialize the basis factor matrices $\bm{U}^{(n)}_{0} \in \mathbb{R}^{I_n \times R_n}$, where $n=\{1, 2, 3\}$. 
Then, we rewrite Eq. (\ref{tucker3}) as follows to solve the $n$-th component matrix $\bm{U}^{(n)}$:
\begin{equation}\label{tucker5}
\begin{aligned}
{\mathop{\argmax}_{{\bm{U}^{(n)}}}}
&\|\bm{U}^{(n)^T}\bm{X} \bar{\times}_{-n}\{\bm{U}^{(n)}\}_{n=1}^3\|_F^2 \\
&s.t. ~~\bm{U}^{(n)^T}\bm{U}^{(n)} = \bm{I}. 
\end{aligned}
\end{equation}

The optimal solution $\bm{U}^{(n)}$ of Eq. (\ref{tucker5}) can be set as the leading $R_n$-dimensional left singular vectors of the matrix $\bm{X} \bar{\times}_{-n}\{\bm{U}^{(n)}\}{_{n=1}^3}$. 
By iteratively updating ${\{{\{\bm{U}^{(n)}_{s}\}}{_{n=1}^3}\}}{_{s=1}^S}$, we can obtain a set of final orthogonal matrices ${\{\bm{U}^{(n)}\}}{_{n=1}^3}$. 
With these final basis factors, the corresponding core tensor $\bm{Y} \in \mathbb{R}^{R_1 \times R_2 \times R_3}$ can be easily solved by Eq. (\ref{tucker1}).

\vspace{1.2mm}
\noindent \textbf{Quantization and de-quantization:} 
Since the core tensor $\bm{Y}$ has both positive and negative values, we take one bit to represent the sign of the original value. 
Let $|\bm{Y}|$ denotes the absolute value of the core tensor. 
With a set of training images, we can easily compute $p(|\bm{Y}|)$, the probability density function (PDF) of the positive core tensor $|\bm{Y}|$. 
The optimal quantizer can be solved as follows by minimizing the quantization error: 
\begin{equation}\label{lloyd}
\begin{aligned}
{Q}^{\star}(|\bm{Y}|) = {\mathop{\argmin}_{\textbf{Q}}}\int p(|\bm{Y}|)({Q}(|\bm{Y}|) - |\bm{Y}|)^2 \,d|\bm{Y}|.
\end{aligned}
\end{equation}

Given a number $M$ of decision intervals, the optimal quantizer is expected to find the set of decision boundaries $\{b_{q}\}_0^M$ and quantized values $\{\hat{Y}_{q}\}{_1^M}$. 
Solving the partial derivative of Eq.(\ref{lloyd}), we could have:
\begin{equation}\label{lloyd1}
\begin{aligned}
\hat{Y}_{q} = \frac{\int_{b_{q-1}}^{b_{q}}|\bm{Y}|p(|\bm{Y}|)\,d|\bm{Y}|}{\int_{b_{q-1}}^{b_{q}}p(|\bm{Y}|)\,d|\bm{Y}|}~~; ~~
b_{q}=\frac{1}{2}(\hat{Y}_{q}+\hat{Y}_{q+1}).
\end{aligned}
\end{equation}
The optimal solutions of Eq.(\ref{lloyd1}) can be easily solved by the Lloyd`s algorithm \cite{lloyd1982least}, outputting the optimal quantizer $Q(\cdot)$ with decision boundaries $\{b_{q}\}{_0^M}$ and quantized values $\{\hat{Y}_{q}\}{_1^M}$.
 
Considering that the range of core tensor values is spatially variant for an input image, we adopt a variable-bits quantization scheme to allocate different quantized bits to the core tensor, which is useful in preserving the major edges and textures. 
More specifically, we first scan the positive core tensor $|\bm{Y}|$ in raster oreder, then utilize the decision boundaries $\{b_{q}\}{_0^M}$ to divide the core tensor into $M$ non-overlapping chunks. 
For each chunk $C_m$ ($m \in [1, M]$), instead of using $\{\hat{Y}_{q}\}_m$ as the quantized values, we define a new quantizer for symbol $|\bm{Y}_i|$ (the $i$-th element of $|\bm{Y}|$):
\begin{equation}\label{quan}
\begin{aligned}
\bar{\bm{Y}}_i = Q(\bm{Y}_i) = [\frac{2^{m}}{C_mMax-C_mMin}](|\bm{Y}_i| - C_mMin),
\end{aligned}
\end{equation}
where $C_mMax$ and $C_mMin$ are the maximum and minimum values of chunk $C_m$, respectively, and $m$ is the number of quantized bits in each chunk $C_m$. 
In this way, each chunk would take $m+1$ bits for the quantization.

Conversely, the de-quantization process can be readily formulated as:
\begin{equation}\label{quan1}
\begin{aligned}
\hat{\bm{Y}}_i= Q^{-1}(\bar{\bm{Y}}_i) = [\frac{\bar{\bm{Y}}_i(C_mMax-C_mMin)}{2^{m}}] + C_mMin.
\end{aligned}
\end{equation}

\vspace{1.2mm}
\noindent \textbf{Tucker reconstruction:} 
With the de-quantized core tensor $\hat{\bm{Y}}$ and the orthogonal matrices $\{\bm{U}^{(n)}\}{_{n=1}^3}$, we can easily obtain an approximation of the original feature data $\hat{\bm{X}}$ by:
\begin{equation}\label{recon}
\begin{aligned}
\hat{\bm{X}} \approx \hat{\bm{Y}} \times_1 \bm{U}^{(1)} \times_2 \bm{U}^{(2)} \times_3  \bm{U}^{(3)}.
\end{aligned}
\end{equation}

The overall TDL is summarized in Algorithm 1. 
With the proposed TDL, we can easily adjust the compression ratio while keeping the size of latent image representation unchanged. 
Finally, we are able to train a single network to achieve LIC with multiple bpp rates.

\begin{table}[t]
\centering
\begin{tabular}{l}
\hline\hline
$\textbf{Algorithm 1:}~\textrm{Tucker Decomposition Layer}$\\
\hline
$\textbf{Input:}$ A $3$-order tensor $\bm{X} \in \mathbb{R}^{I_1 \times I_2 \times I_3}$,\\
~~~~~~~~~the desired output rank-$(R_1, R_2, R_3)$,\\
~~~~~~~~~a number of decision intervals $M$.\\
$\textbf{Output:}$ An approximation of original data $\hat{\bm{X}} \in \mathbb{R}^{I_1 \times I_2 \times I_3}$,\\
~~~~~~~~~~~a $3$-order quantized core tensor $\bar{\bm{Y}} \in \mathbb{R}^{R_1 \times R_2 \times R_3}$,\\ 
~~~~~~~~~~~the orthogonal matrices $\{\bm{U}^{(n)} \in \mathbb{R}^{I_n \times R_n}\}_{n=1}^3$.\\ 
$\bm{1:}$ $\bm{U}_0^{(n)}\leftarrow R_n$leading left singular vectors of $\bm{X}_{(n)}$, for $n = 1, 2, 3$; \\
$\bm{2:}$ \textbf{For} $s = 0, 1, 2,$ ... (until converged), \textbf{do}:\\
$\bm{3:}$ ~~~~~\textbf{For} $n = 1, 2, 3$, \textbf{do}:\\
$\bm{4:}$ ~~~~~~~~~~$\bm{U}_{s+1}^{(n)} \leftarrow \bm{U}_{s}^{(n)}$ by Eq. (\ref{tucker5});\\
$\bm{5:}$ ~~~~~\textbf{End for};\\
$\bm{6:}$ \textbf{End for};\\
$\bm{7:}$ Let $\{\bm{U}\}$=$\{\bm{U}_S\}$, where $S$ is the index of the final result of step 2;\\
$\bm{8:}$ Compute decision boundaries $\{b_{q}\}_0^M$ by Eq.(\ref{lloyd1});\\  
$\bm{9:}$ Divide $|\bm{Y}|$ into $M$ non-overlapping chunks;\\
$\bm{10:}$ $\bar{\bm{Y}} \leftarrow Q(|\bm{Y}|)$ by Eq. (\ref{quan});\\
$\bm{11:}$ $\hat{\bm{Y}} \leftarrow Q^{-1}(\bar{\bm{Y}})$ by Eq. (\ref{quan1});\\
$\bm{12:}$ $\hat{\bm{X}} \leftarrow \hat{\bm{Y}}, \bm{U}^{(1)}, \bm{U}^{(2)}, \bm{U}^{(3)}$ by Eq. (\ref{recon});\\
$\bm{13:}$ return $\hat{\bm{X}}, \{\bm{U}^{(n)}\}_{n=1}^3, \bar{\bm{Y}}$.\\
\hline\hline
\end{tabular}
\end{table}

\begin{table}[t]
\centering
\begin{tabular}{l}
\hline\hline
$\textbf{Algorithm 2:}~\textrm{All-in-One Training}$\\
\hline
$\textbf{Input:}$ A set of training data $\{\bm{x}_i\}_{i=1}^N$,\\
~~~~~~~~~$G$ groups of desired ranks $\{R_1, R_2, R_3\}_{g=1}^G$,\\
~~~~~~~~~$G$ groups of decision intervals $\{M\}_{g=1}^G$.\\
$\textbf{Output:}$ The optimal parameters $\{\Omega, \Theta, \Pi\}$.\\ 
$\bm{1:}$ $\{\Omega, \Theta, \Pi\}_0 \leftarrow$ train encoder-decoder with Eq. (\ref{l2}) / (\ref{msssim});\\
$\bm{2:}$ \textbf{For} $k = 1, 2, 3,$ ... (until converged), \textbf{do}:\\
$\bm{3:}$ ~~~${\{\{\bm{z}_i\}_{i=1}^N\}}_k = E(\{\bm{x}_i\}_{i=1}^N, \Omega_{k-1})$;\\
$\bm{4:}$ ~~~${\{\{\bm{Y}_i\}_{i=1}^N\}}_k \leftarrow$ decompose ${\{\{\bm{z}_i\}_{i=1}^N\}}_k$ by Algorithm 1;\\
$\bm{5:}$ ~~~\textbf{For} $g = 1, 2, 3, \cdots, G$, \textbf{do}:\\
$\bm{6:}$ ~~~~~~${\{\{\{b_{q}\}_0^M\}_g\}}_k \leftarrow$ compute the $g$-th group decision boundaries\\
~~~~~~~~~~~~~~~~~~~~~~~~~~~~~~~~~~~~~~~~~~~~~~~~~~~~~~~~~~~~~~~~~~~~~~by Eq. (\ref{lloyd1});\\
$\bm{7:}$ ~~~\textbf{End for};\\
$\bm{8:}$ ~~~Fixed the TDL;\\
$\bm{9:}$ ~~~\textbf{For} $epoch = 1, 2, 3,$ ... (until converged), \textbf{do}:\\
$\bm{10:}$ ~~~~$\hat{g}$ = epoch $\bmod$ G;\\
$\bm{11:}$ ~~~~${\{\{\Omega, \Theta, \Pi\}_{epoch+1}\}}_k \leftarrow$ use ${\{\{R_1, R_2, R_3, \{b_{q}\}_0^M\}_{(\hat{g}+1)}\}}_k$\\ 
~~~~~~~~~~~~~~~~~~~~~~~~and Eq. (\ref{l2}) / (\ref{msssim}) to update encoder-TDL-decoder;\\
$\bm{12:}$ ~~\textbf{End for};\\
$\bm{13:}$ ~~Let ${\{\Omega, \Theta, \Pi\}}_k$ = ${\{\{\Omega, \Theta, \Pi\}_{E}\}}_k$,\\
~~~~~~~~~~~~~~~~~~~~~~~~where $E$ is the index of the final result of step 9;\\ 
$\bm{14:}$ \textbf{End for};\\
$\bm{15:}$ Let $\{\Omega, \Theta, \Pi\}$ = $\{\Omega, \Theta, \Pi\}_K$,\\
~~~~~~~~~~~~~~~~~~~~~~~~where $K$ is the index of the final result of step 2;\\ 
$\bm{16:}$ $\{\{\hat{\bm{z}}_i\}_{i=1}^N\}_{g=1}^G$ = $T^{-1}(T(E(\{\bm{x}_i\}_{i=1}^N, \Omega)))$ by Algorithm 1;\\  
$\bm{17:}$ $\{\Theta, \Pi\} \leftarrow$ update the decoder network with \\ 
~~~~~~~~~~~~~~~~~~~~~Eq. (\ref{l2}) / (\ref{msssim}) and $\{\{\hat{\bm{z}}_i\}_{i=1}^N\}_{g=1}^G$;\\
$\bm{18:}$ return $\{\Omega, \Theta, \Pi\}$.\\
\hline\hline
\end{tabular}
\end{table}

\section{All-in-One Training}
\label{sec:all}
Instead of training a specific network for a certain compression ratio as in previous deep LIC methods \cite{toderici2015variable, toderici2017full, balle2017end, theis2017lossy, li2017learning, rippel2017real, mentzer2018conditional, johnston2017improved, agustsson2017soft}, the proposed TDNet allows an all-in-one training strategy to enable a single network to compress an image at multiple compression ratio. 
More specifically, we first train an encoder-decoder network without the TDL to learn some initial parameters ${\{\Omega, \Theta, \Pi\}}_0$. 
We can then calculate a set of latent image representations of the input images $\{\bm{x}_i\}{_{i=1}^N}$ by $\{{\bm{z}_i\}}{_{i=1}^N} = E(\{\bm{x}_i\}{_{i=1}^N}, {\Omega}_0)$. 
Given $G$ groups of desired output ranks and quantization levels $\{R_1, R_2, R_3, M\}{_{g=1}^G}$ and the obtained latent image representations $\{{\bm{z}_i\}}{_{i=1}^N}$, we can use the proposed TDL to calculate $G$ groups of decision boundaries ${\{\{b_{q}\}_0^M\}}{_{g=1}^G}$. 
The TDL can then be initialized after obtaining the decision boundaries $\{b_{q}\}{_0^M}$.

With the initialized TDL, we can use Eq.(\ref{l2}) or Eq. (\ref{msssim}) to jointly fine-tune the encoder-TDL-decoder network by minimizing the loss function. 
The latent image representations at multiple bpp rates will be taken into consideration during the training process. 
In each training epoch, we first decide which group of ranks and decision boundaries will be used by calculating $\hat{g} = mod\text{(epoch, G)}$, then take this group of desired output ranks and decision boundaries ${\{R_1, R_2, R_3,\{b_{q}\}_0^M\}}_{(\hat{g}+1)}$ to update the TDL and fine-tune the parameters $\{\Omega, \Theta, \Pi\}$ of the whole network. 
To obtain $G$ groups of optimal decision boundaries ${\{\{b_{q}\}_0^M\}}{_{g=1}^G}$ and network parameters $\{\Omega, \Theta, \Pi\}$, an iterative training scheme can be used, \ie, fix the encoder-decoder network to update the TDL decision boundaries ${\{\{b_{q}\}_0^M\}}{_{g=1}^G}$ by solving Eq.(\ref{lloyd1}), and fix the TDL to update the network parameters $\{\Omega, \Theta, \Pi\}$. 
Such an alternative optimization process continues till the loss function in Eq.(\ref{l2}) or Eq. (\ref{msssim}) converges.

After the TDNet converges, we can use the optimal decision boundaries ${\{\{b_{q}\}_0^M\}}{_{g=1}^G}$ and network parameters $\{\Omega, \Theta, \Pi\}$ to compress and reconstruct images with different bpp rates.
The overall all-in-one training scheme is summarized as Algorithm 2.

\begin{figure*}
\footnotesize
\centering
\subfigure{
\begin{minipage}[t]{0.33\textwidth}
\centering
\includegraphics[width=1\textwidth]{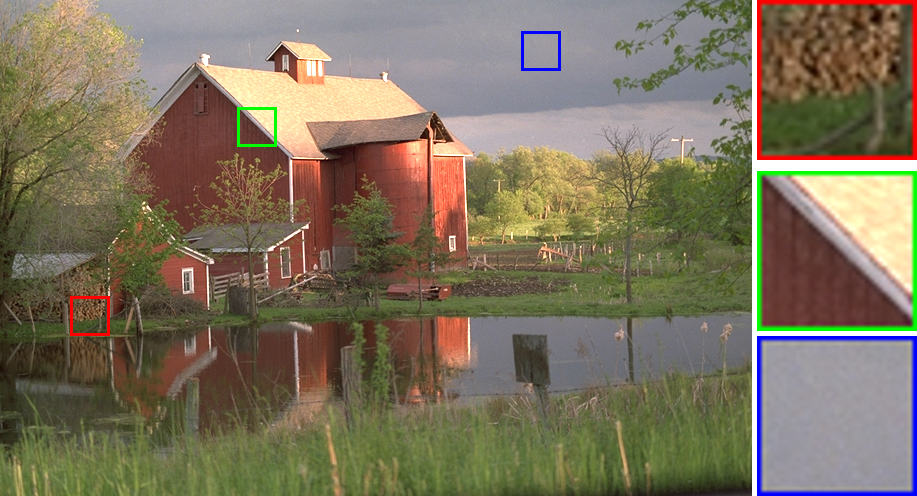}
{(a) \textbf{Original}}
\end{minipage}
\begin{minipage}[t]{0.33\textwidth}
\centering
\includegraphics[width=1\textwidth]{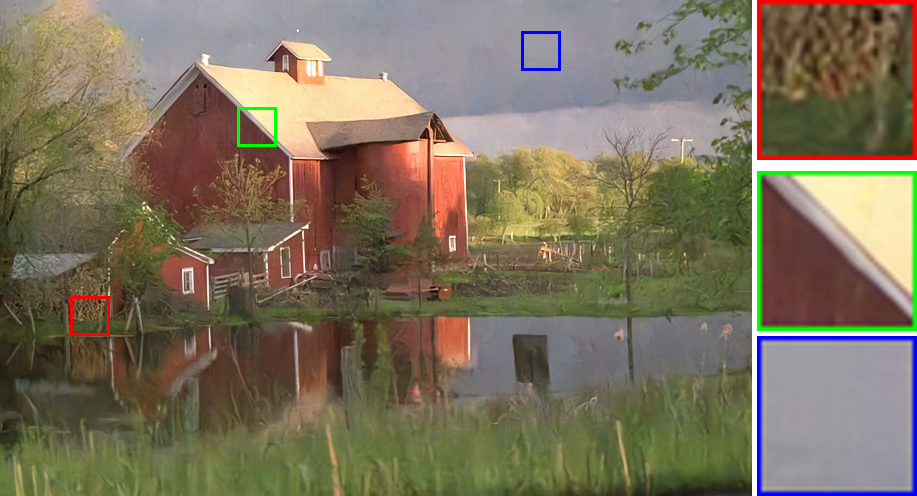}
{(b) \textbf{Single network} (BPP: 0.352 PSNR: 30.36)}
\end{minipage}
\begin{minipage}[t]{0.33\textwidth}
\centering
\includegraphics[width=1\textwidth]{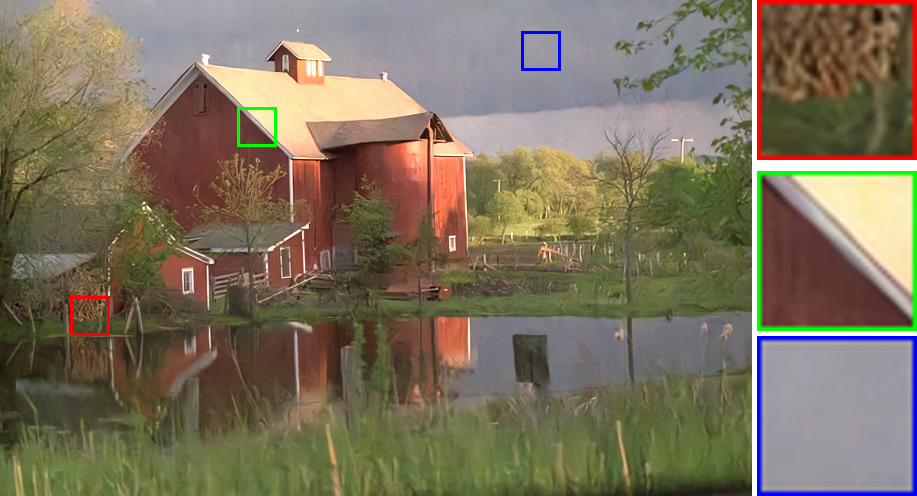}
{(c) \textbf{Individual network} (BPP: 0.348 PSNR: 30.45)}
\end{minipage}
}
\caption{Visual comparison on image ``$house$" by the proposed single network and individual network.}
\label{fig:as2}
\end{figure*}

\section{Experimental Results}
\label{sec:expri}
In this section, we first present the experimental settings, including training and testing datasets, as well as parameter settings. 
We then discuss the performance of TDNet using a single network and multiple networks.
Finally, we compare TDNet with state-of-the-art LIC methods.

\subsection{Experimental Settings}
\noindent \textbf{Datasets:}
It is generally agreed that a larger scale training dataset which covers various image contents and structures will bring benefit to train a stable deep LIC network. 
Therefore, we mix the MS COCO test2017 dataset \cite{lin2014microsoft}, the DIV2K dataset \cite{Agustsson} and the Waterloo Exploration dataset \cite{ma2017waterloo} together as the training dataset. 
The MS COCO test2017 dataset contains $40,652$ images which cover a great diversity of objects and scenes. 
The DIV2K dataset has $900$ high-resolution images with complex structures and texture patterns. 
The Waterloo Exploration dataset contains $4,744$ elaborately selected high quality natural images. 
We first crop these images into $320\times320$ patches\footnote{We experimentally found that using the same network architecture, the larger the training patches are, the better the results would be. To trade off between GPU memory and compression results, we set the size of training patches as $320\times320$.}, then randomly flip them. 
With around $100,000$ image patches, we could train a robust model for LIC with multiple bpp rates.

For the testing, we use two different test datasets for comprehensive evaluation: the Kodak PhotoCD dataset \footnote{\url{http://r0k.us/graphics/kodak/}}, which contains 24 natural images, and the McMaster dataset \cite{zhang2011color}, which contains 18 high quality images. 
Note that all those images are widely used for the evaluation of image processing methods and they are not included in the training dataset.

\vspace{1.2mm}
\noindent \textbf{Parameter Settings:}
In the training phase, we set $G=4$ groups of desired ranks and decision intervals ${\{R_1, R_2, R_3, M\}}_{g=1}^4$ to train the TDNet: $\{\{38, 37, 28, 5\},$ $\{36, 35, 26, 4\},$ $\{34, 31, 23, 3\},$ $\{34, 30, 22, 3\}\}$. 
The mini-batch size is set to $7$. 
We initialize the network weights by the method in \cite{he2015delving} and adopt the Adam solver \cite{kingma2014adam} to optimize the network parameters $\{\Omega, \Theta, \Pi\}$. 
The learning rate starts from 1$e-$4 and is then fixed to 1$e-$5 when the training error stops decreasing.
The training is terminated when the training error does not decrease in $20$ sequential epochs. 
For the other hyper-parameters of Adam, we utilize the default setting. 
We employ the context-based adaptive binary arithmetic coding (CABAC) \cite{marpe2003context} for lossless entropy coding.

We experimentally found that the alternative optimization process of our TDNet compressor (refer to Algorithm 2 please) will converge in less than $k=4$ iterations. 
The parameter $\lambda$ in our loss function Eq. (\ref{l2}) or Eq. (\ref{msssim}) is set to $0.4$ by experience. 
The network is trained in CAFFE \cite{jia2014caffe} with an Nvidia Titan Xp GPU. In our PC with Intel(R) Core(TM) i9-7900X CPU @ 3.3GHz, 96G RAM, the training process costs about 3 days. 

Though we use $4$ groups of ranks and decision intervals $\{R_1, R_2, R_3, M\}$ to train the TDNet, to validate the generality of the trained network, in the testing phase we use $6$ groups of ranks and decision intervals to test the performance of TDNet: $\{\{38, 37, 28, 5\},$ $\{36, 35, 26, 4\},$ $\{35, 32, 23, 4\},$ $\{34, 31, 23, 3\},$ $\{34, 30, 22, 3\}\}$ and $\{34, 30, 22, 2\}\}$. 
As we will see in the following sections, our TDNet achieves highly competitive performance.

\begin{figure}
\centering
\includegraphics[width=0.5\textwidth]{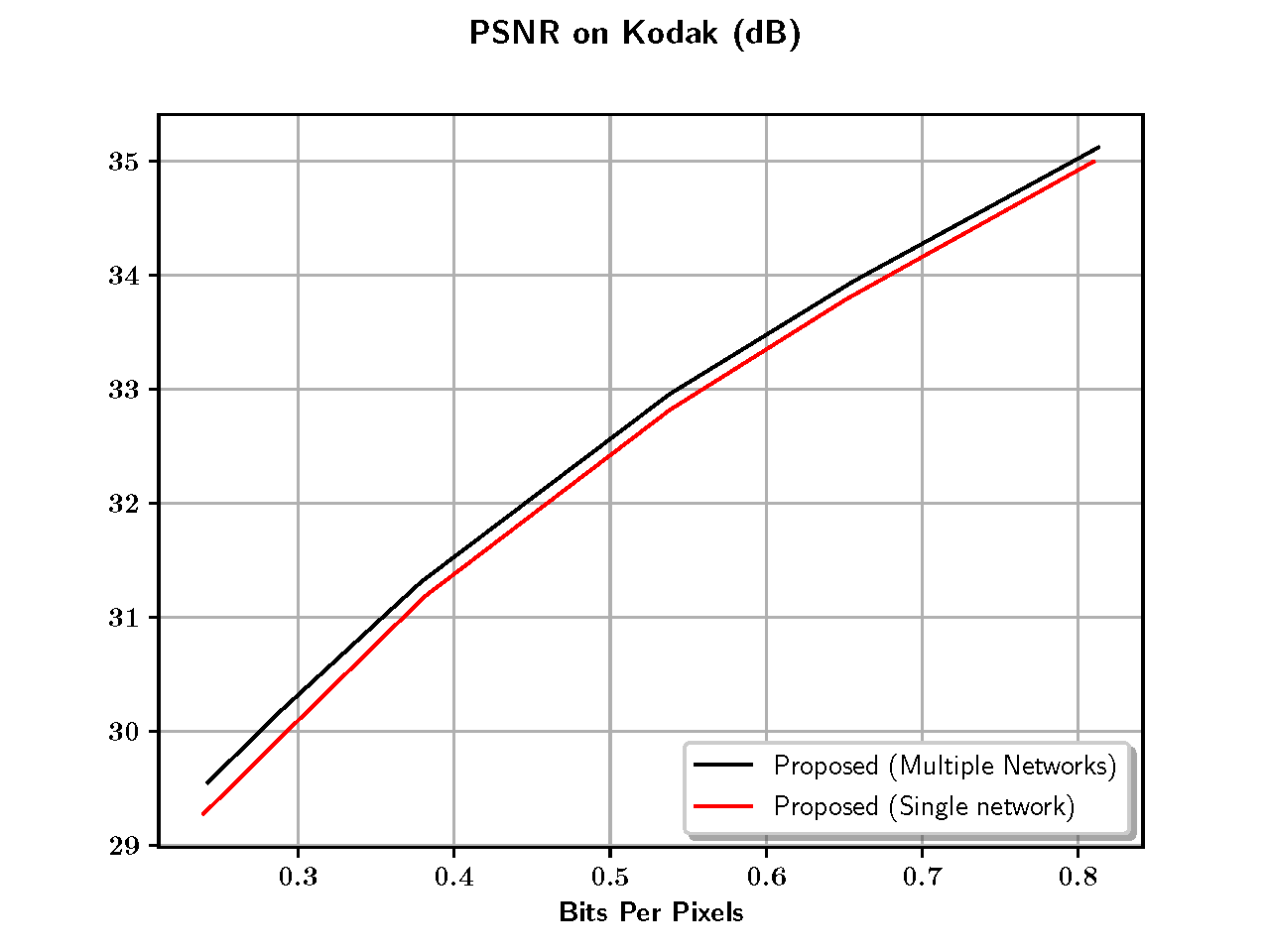}
\caption{Rate-distortion curves by the single network and multiple networks on the Kodak dataset.}
\label{fig:as}
\end{figure}

\begin{figure*}
\footnotesize
\centering
\subfigure{
\begin{minipage}[t]{0.5\textwidth}
\centering
\includegraphics[width=1\textwidth]{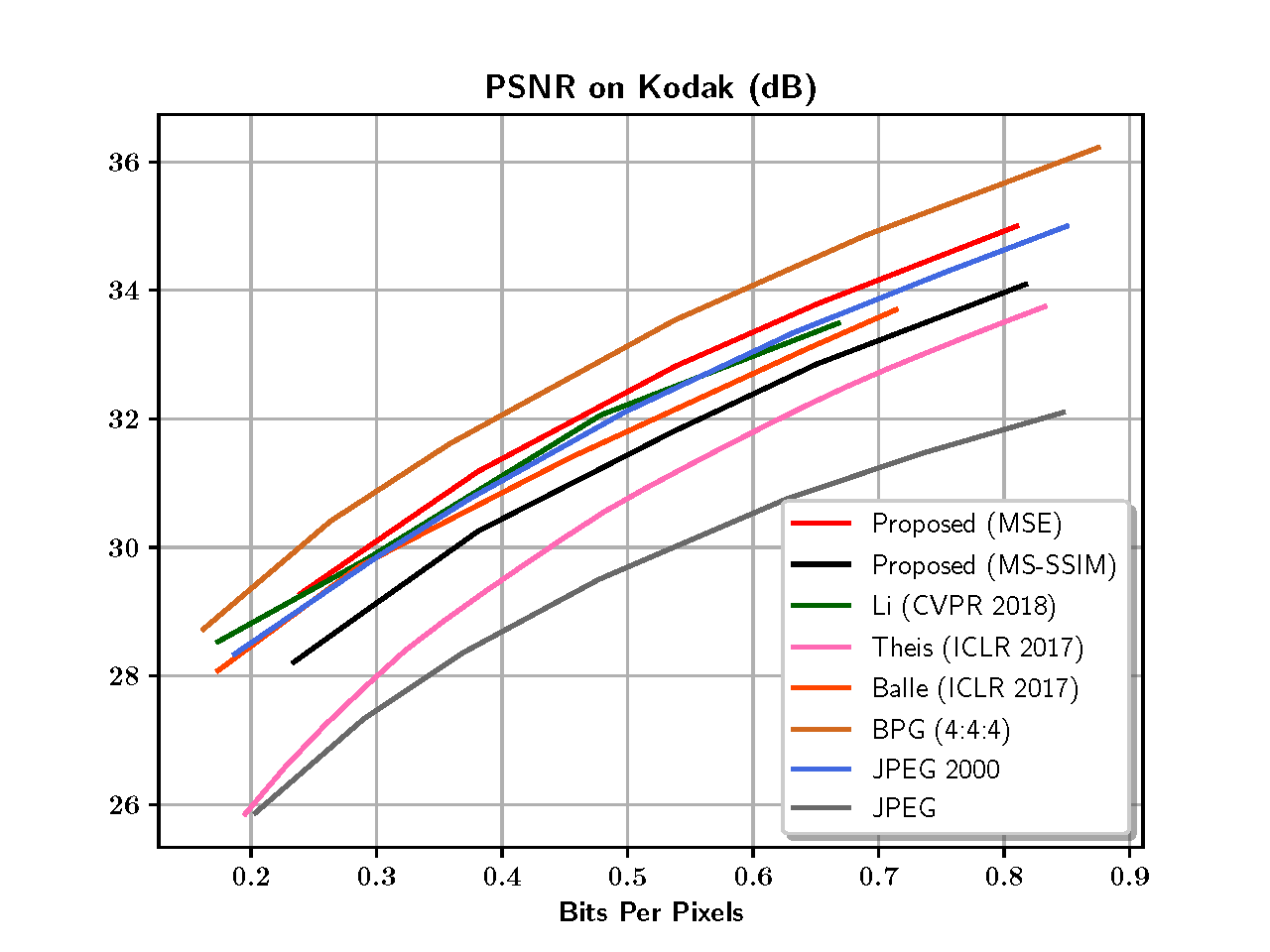}
{(a)}
\end{minipage}
\begin{minipage}[t]{0.5\textwidth}
\centering
\includegraphics[width=1\textwidth]{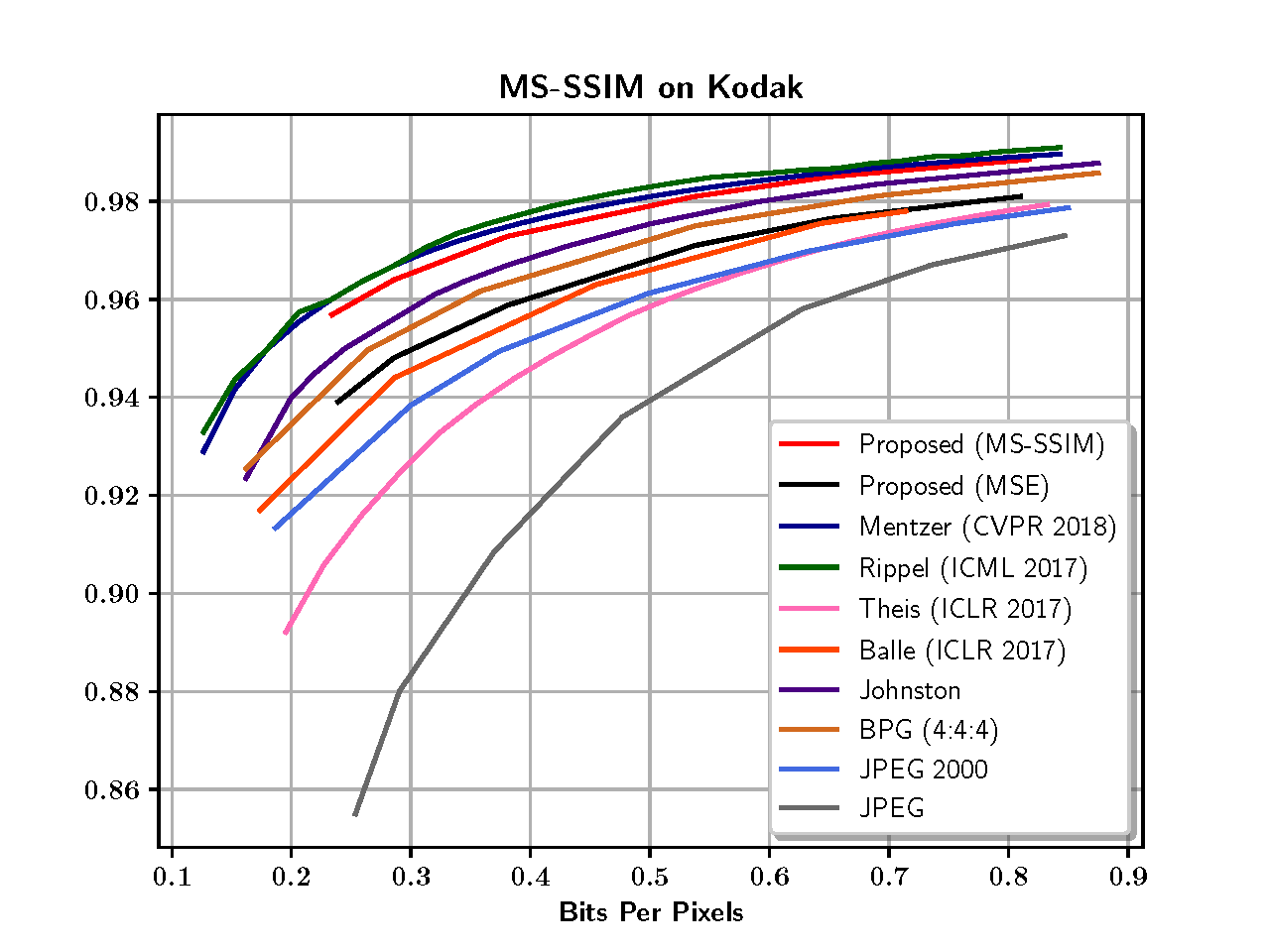}
{(b)}
\end{minipage}
}
\subfigure{
\begin{minipage}[t]{0.5\textwidth}
\centering
\includegraphics[width=1\textwidth]{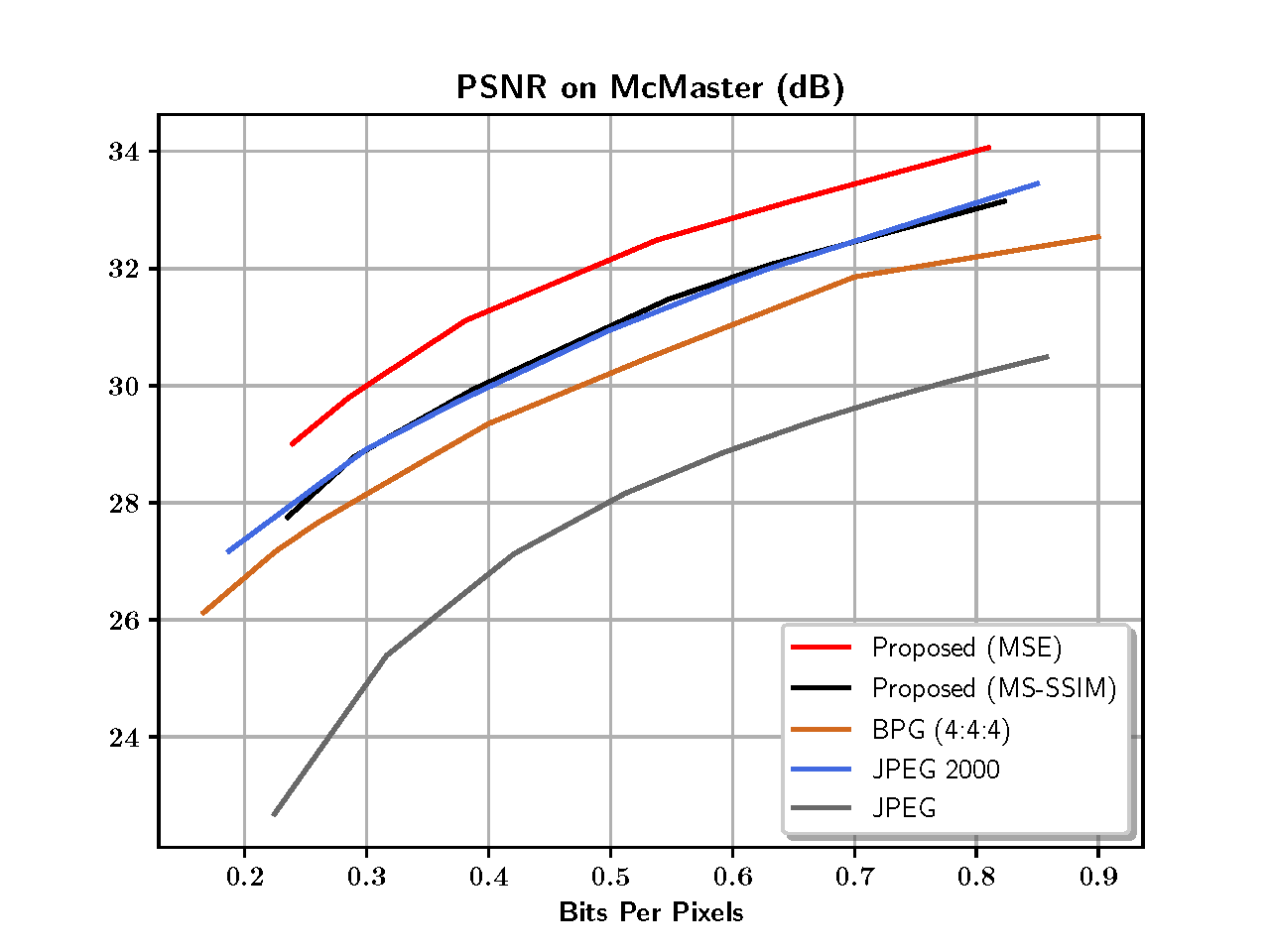}
{(c)}
\end{minipage}
\begin{minipage}[t]{0.5\textwidth}
\centering
\includegraphics[width=1\textwidth]{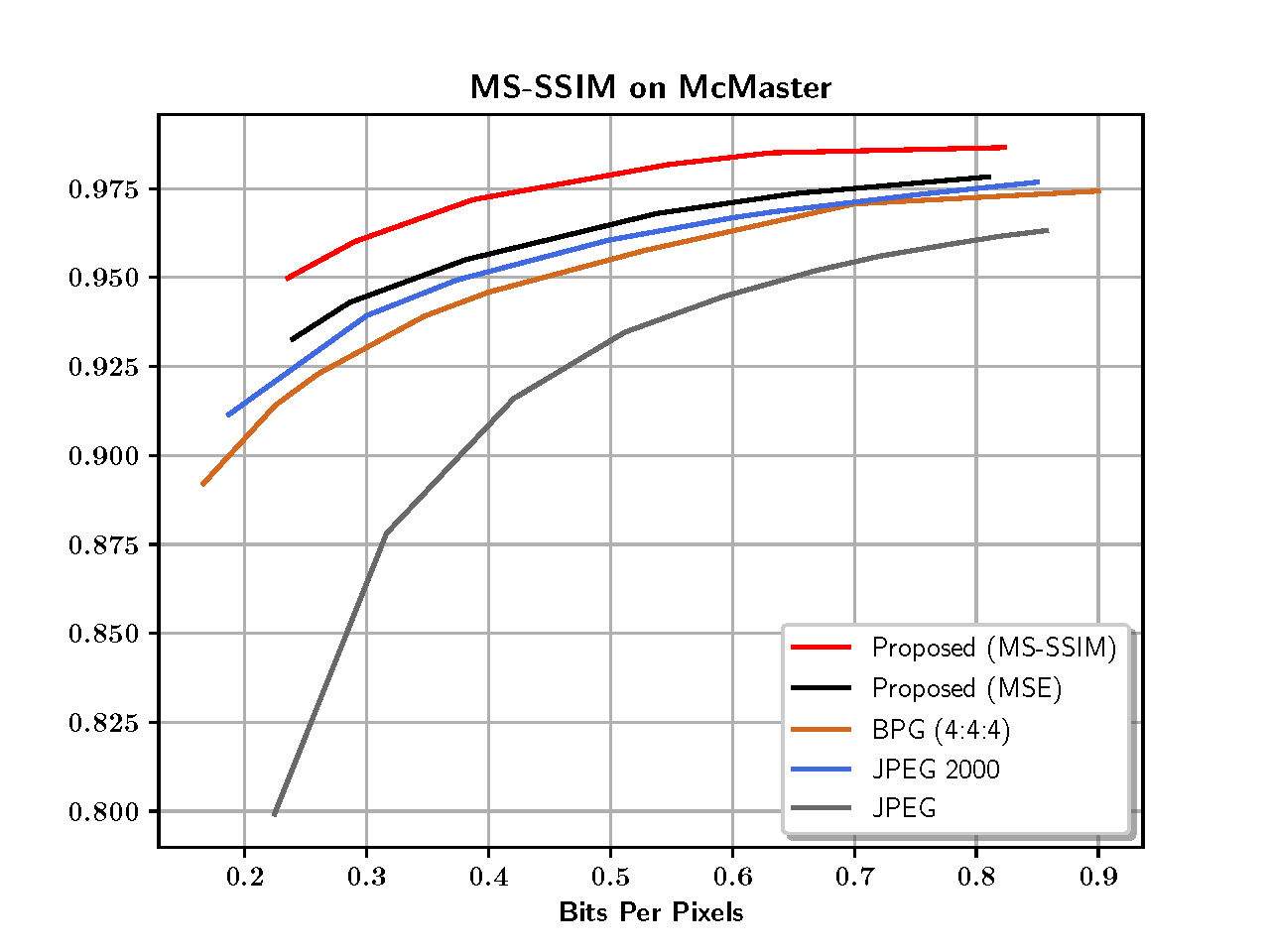}
{(d)}
\end{minipage}
}
\caption{Comparison of the rate-distortion curves on Kodak and McMaster datasets.}
\label{fig:cur}
\end{figure*}

\subsection{Single Network vs. Multiple Networks}
The proposed TDNet allows an "all-in-one" training strategy to learn a single network to perform LIC at multiple bpp rates. 
To validate the effectiveness of our "single network multiple bpp" scheme, in this section we also train six individual TDNet compressors for the six groups of ${\{R_1, R_2, R_3, M\}}_{g=1}^6$  (i.e., six bpp rates), respectively, and compare the performance of single TDNet and multiple TDNet at different bpp rates.

Figure \ref{fig:as2} compares the visual quality of compressed images $house$ by those two training strategies at around $0.35$bpp. 
We also show the zoom-in images of a smooth background area, a texture area and a large edge area. 
It can be seen that both the two training strategies produce good preservation of image edges and details, and they have very small visual difference.

Figure \ref{fig:as} shows the PSNR based rate-distortion curves by the single network and multiple networks on the Kodak dataset. 
Note that the six points on the curve of multiple networks is obtained by a TDNet trained at a specific bpp. 
As one can see, the rate-distortion curve of the trained single TDNet is very close to the curve obtained by multiple networks. On average, its PSNR is only $0.181$dB lower than that of multiple networks.

\begin{figure*}
\footnotesize
\centering
\subfigure{
\begin{minipage}[t]{0.33\textwidth}
\centering
\includegraphics[width=1\textwidth]{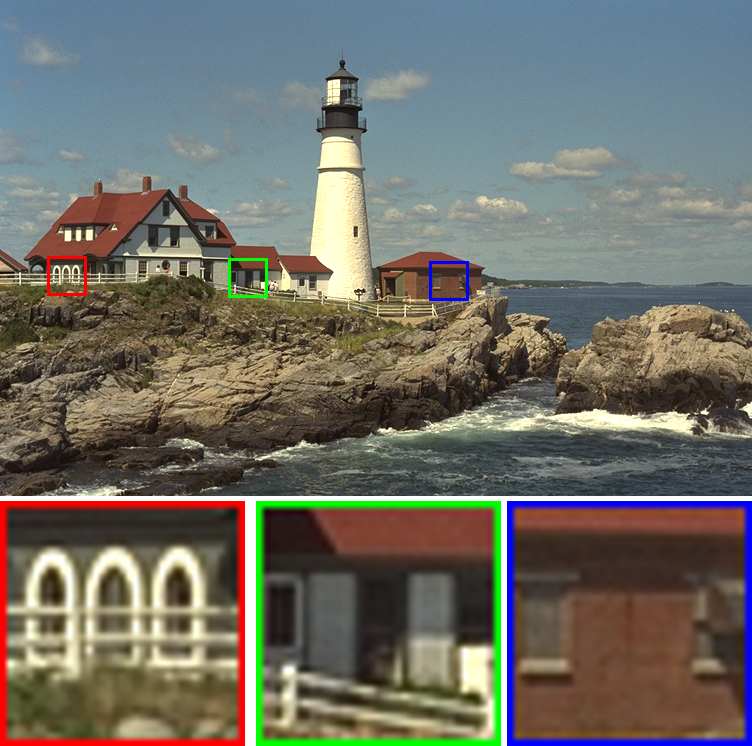}
{(a) \textbf{Original}}
\end{minipage}
\begin{minipage}[t]{0.33\textwidth}
\centering
\includegraphics[width=1\textwidth]{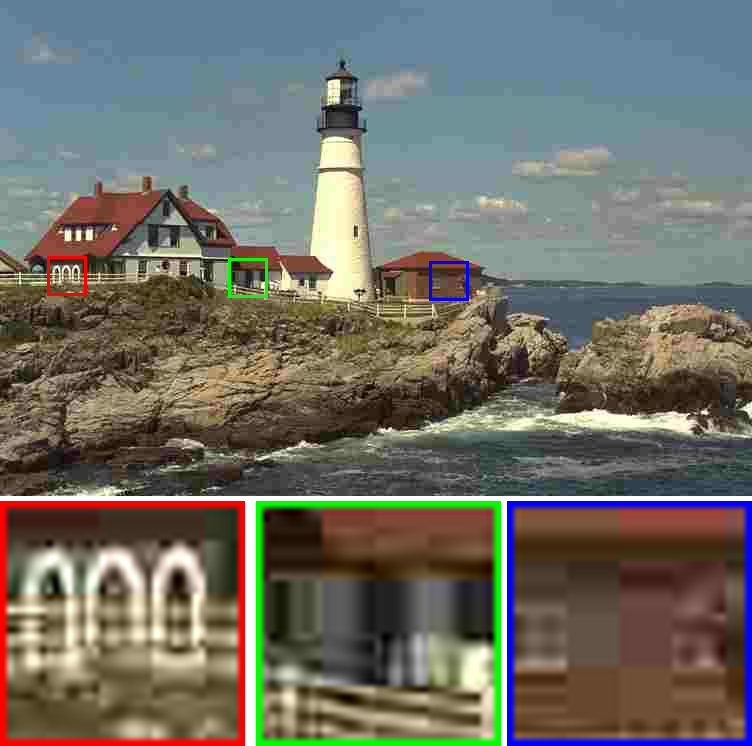}
{(b) \textbf{JPEG} ~~~BPP: 0.308} 
\end{minipage}
\begin{minipage}[t]{0.33\textwidth}
\centering
\includegraphics[width=1\textwidth]{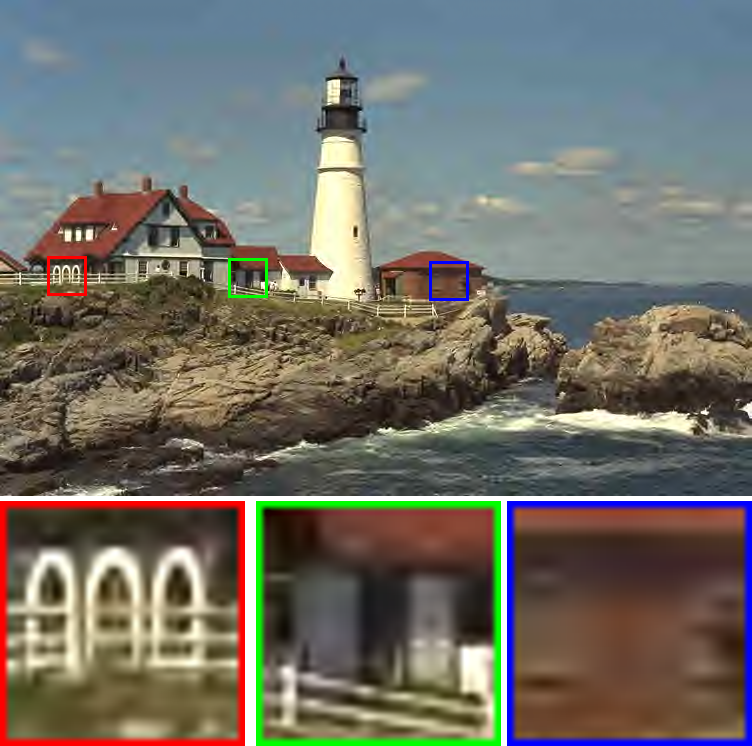}
{(c) \textbf{JPEG 2000} ~~~BPP: 0.279}
\end{minipage}
}
\subfigure{
\begin{minipage}[t]{0.33\textwidth}
\centering
\includegraphics[width=1\textwidth]{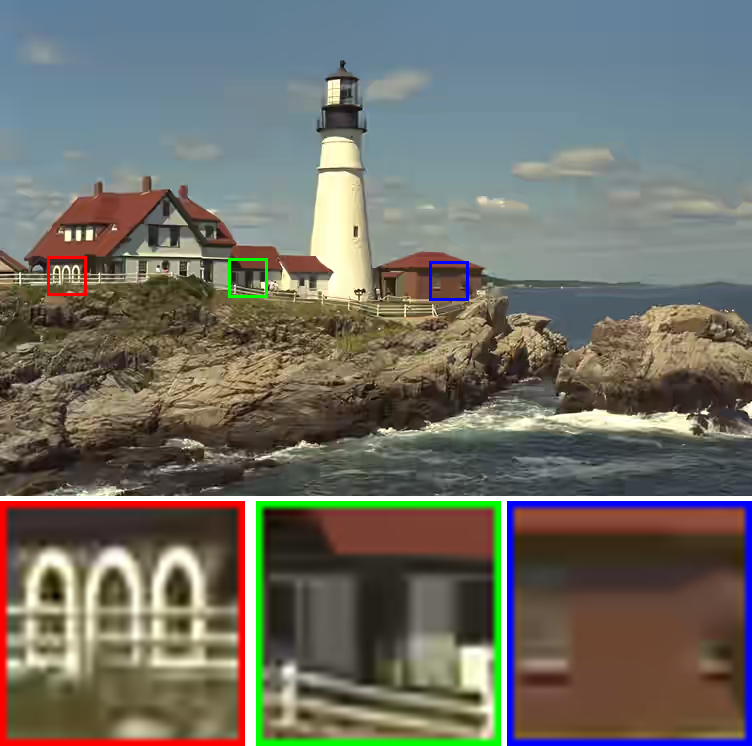}
{(d) \textbf{BPG} ~~~BPP: 0.264} 
\end{minipage}
\begin{minipage}[t]{0.33\textwidth}
\centering
\includegraphics[width=1\textwidth]{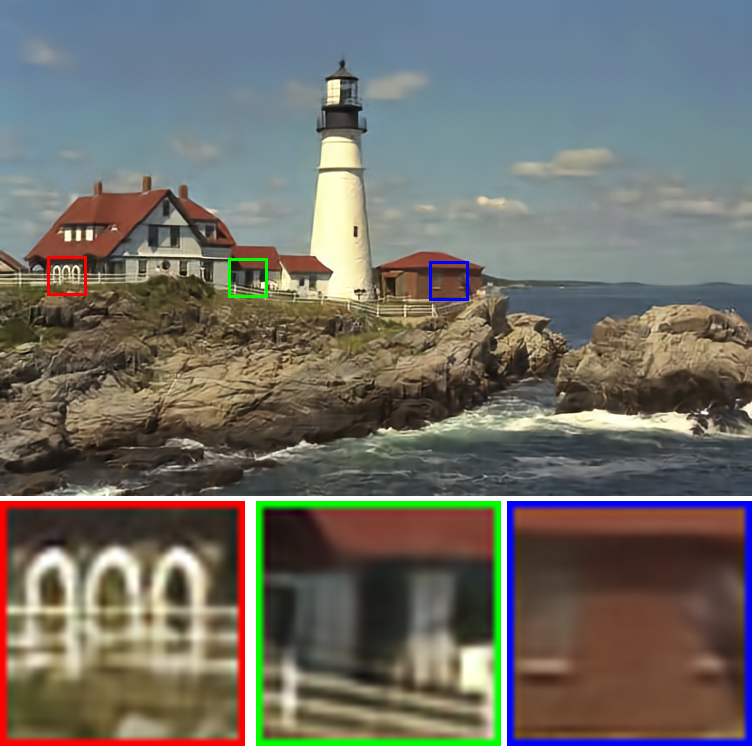}
{(e) \textbf{Ball{\'e}} ~~~BPP: 0.288} 
\end{minipage}
\begin{minipage}[t]{0.33\textwidth}
\centering
\includegraphics[width=1\textwidth]{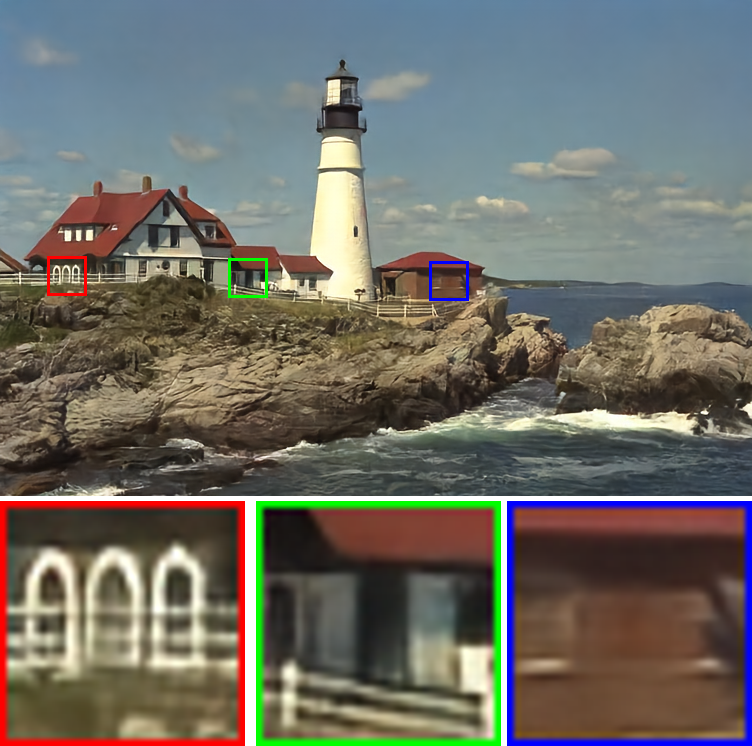}
{(f) \textbf{Li} ~~~BPP: 0.301} 
\end{minipage}
}
\subfigure{
\begin{minipage}[t]{0.33\textwidth}
\centering
\includegraphics[width=1\textwidth]{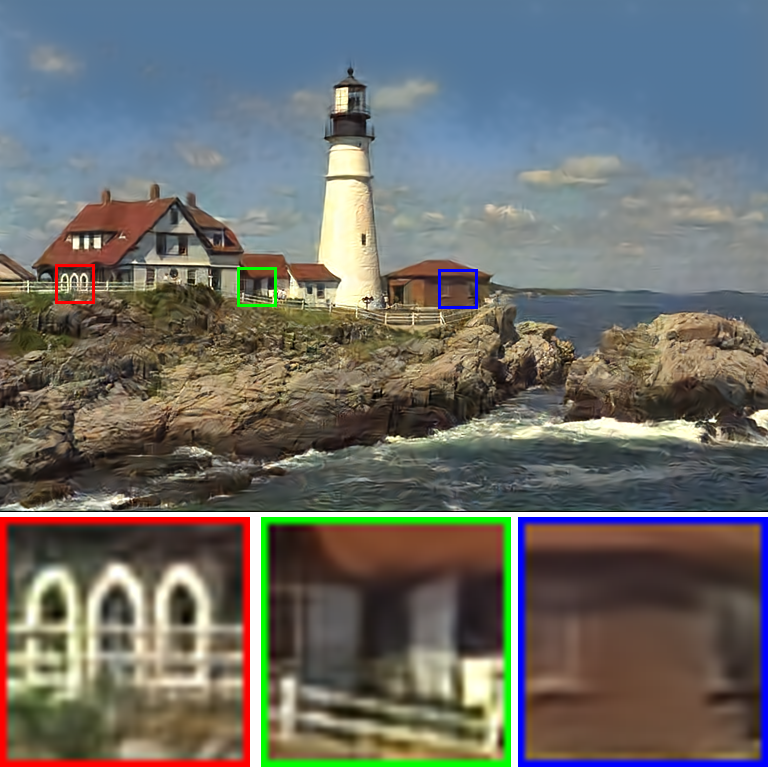}
{(g) \textbf{Theis} ~~~BPP: 0.375} 
\end{minipage}
\begin{minipage}[t]{0.33\textwidth}
\centering
\includegraphics[width=1\textwidth]{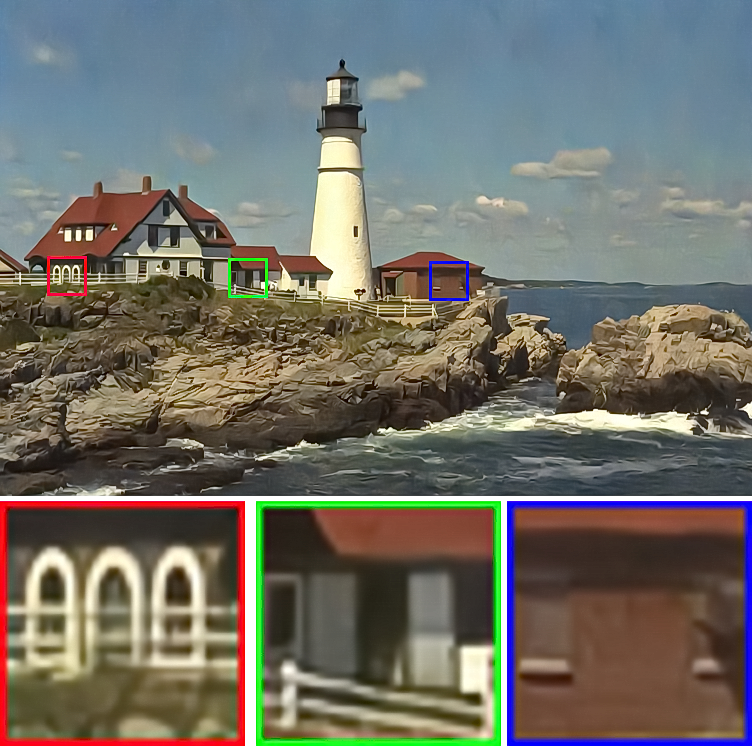}
{(h) \textbf{Proposed (MS-SSIM)} ~~~BPP: 0.301} 
\end{minipage}
\begin{minipage}[t]{0.33\textwidth}
\centering
\includegraphics[width=1\textwidth]{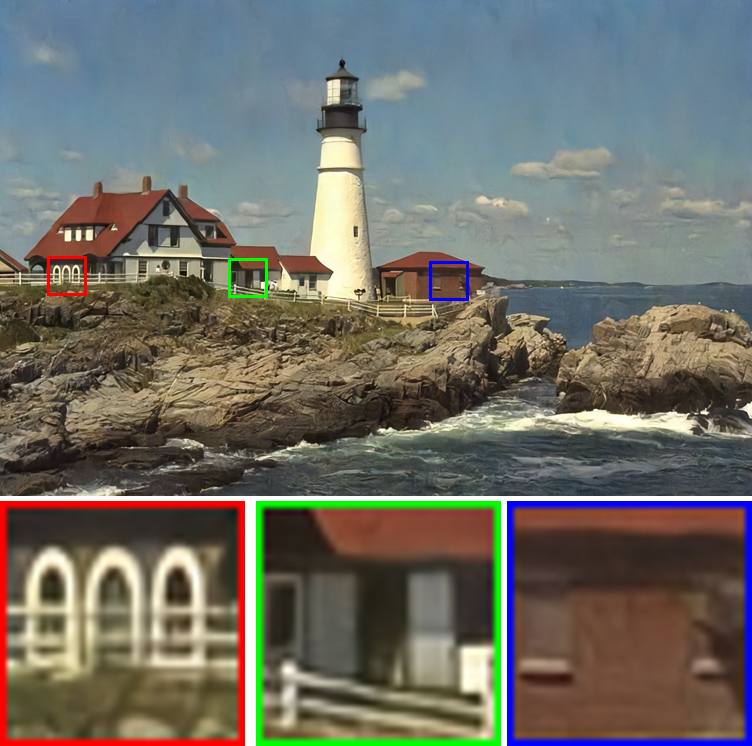}
{(i) \textbf{Proposed (MSE)} ~~~BPP: 0.296} 
\end{minipage}
}
\caption{Visual comparison on image ``$lighthouse$" by different methods at a compression rate around 0.3bpp (Image from Kodak dataset)}
\label{fig:re1}
\end{figure*}
\begin{figure*}
\footnotesize
\centering
\subfigure{
\begin{minipage}[t]{0.33\textwidth}
\centering
\includegraphics[width=1\textwidth]{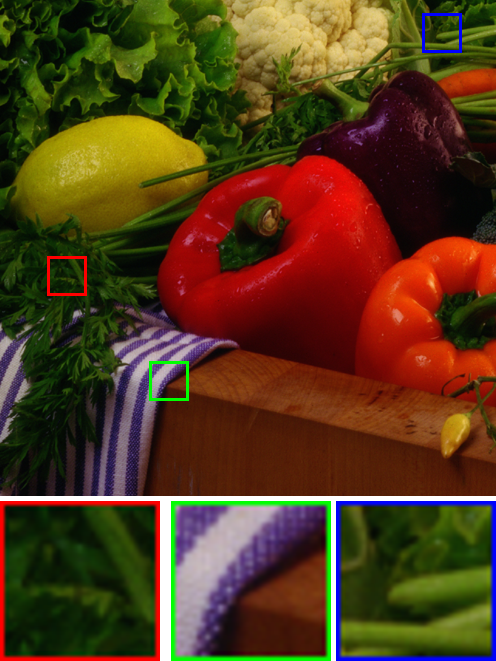}
{(a) \textbf{Original}}
\end{minipage}
\begin{minipage}[t]{0.33\textwidth}
\centering
\includegraphics[width=1\textwidth]{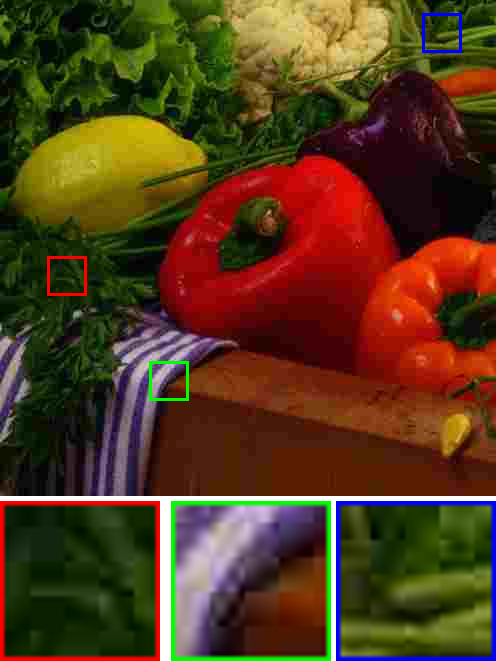}
{(b) \textbf{JPEG} ~~~BPP: 0.339} 
\end{minipage}
\begin{minipage}[t]{0.33\textwidth}
\centering
\includegraphics[width=1\textwidth]{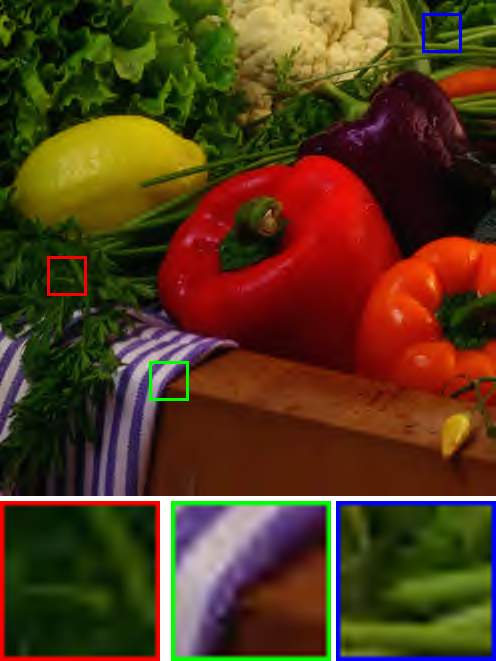}
{(c) \textbf{JPEG 2000} ~~~~BPP: 0.324} 
\end{minipage}
}
\subfigure{
\begin{minipage}[t]{0.33\textwidth}
\centering
\includegraphics[width=1\textwidth]{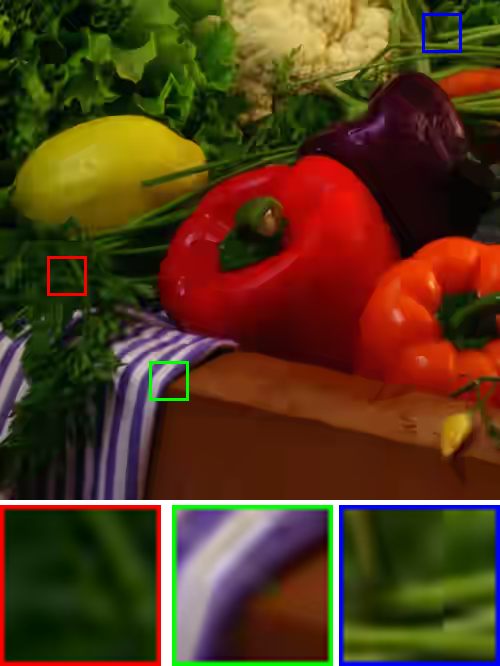}
{(d) \textbf{BPG} ~~~BPP: 0.294} 
\end{minipage}
\begin{minipage}[t]{0.33\textwidth}
\centering
\includegraphics[width=1\textwidth]{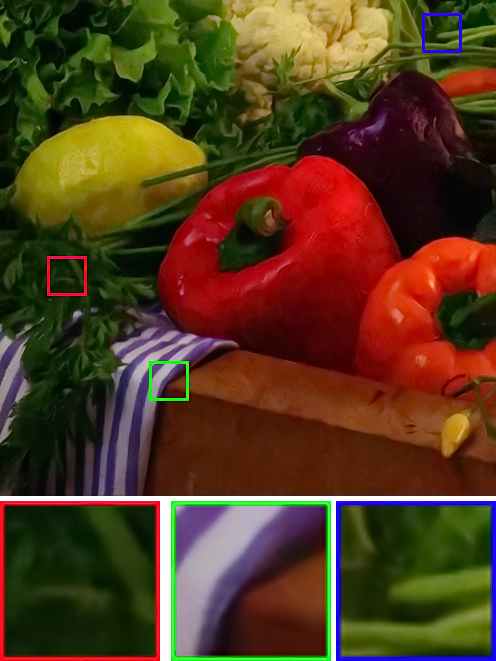}
{(e) \textbf{Proposed (MS-SSIM)} ~~~~~BPP: 0.296} 
\end{minipage}
\begin{minipage}[t]{0.33\textwidth}
\centering
\includegraphics[width=1\textwidth]{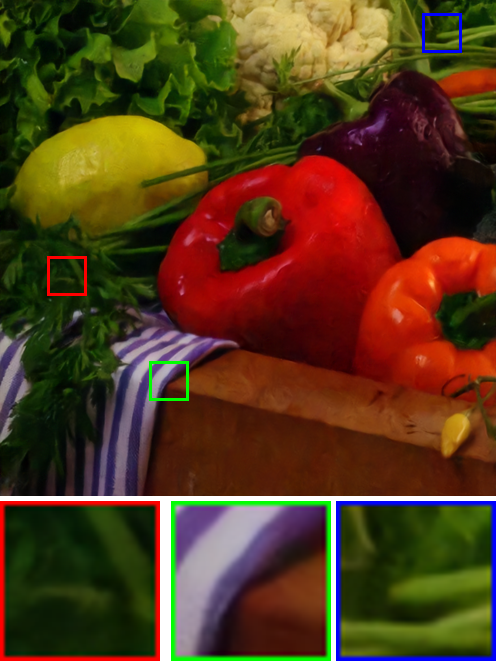}
{(f) \textbf{Proposed (MSE)} ~~~~BPP: 0.291} 
\end{minipage}
}
\caption{Visual comparison on image ``$peppers$" by different methods at a compression rate around 0.3bpp (Image from McMaster dataset)}
\label{fig:re2}
\end{figure*}

\subsection{Results}
We compare our proposed TDNet with both traditional LIC Codecs and CNN-based LIC methods.

\vspace{1.2mm}
\noindent \textbf{Traditional LIC Codecs:}
The compared traditional LIC codecs include JPEG (implemented by libjpeg\footnote{http://libjpeg.sourceforge.net/}), JPEG 2000 (implemented by Matlab) and the state-of-the-art compression format better portable graphics (BPG)\footnote{5https://bellard.org/bpg/}. 
Following \cite{rippel2017real}, we use BPG with the setting of $4:4:4$ chroma format.

\vspace{1.2mm}
\noindent \textbf{Deep LIC Methods:}
The compared CNN-based LIC methods include Ball{\'e} \etal \cite{balle2017end}\footnote{\url{http://www.cns.nyu.edu/lcv/iclr2017/}}, Theis \etal \cite{theis2017lossy}\footnote{\url{http://theis.io/compressive_autoencoder/}}, Li \etal \cite{li2017learning}\footnote{\url{http://www2.comp.polyu.edu.hk/~15903062r/index.html}}, Johnston \etal \cite{johnston2017improved}, Rippel $\&$ Bourdev \cite{rippel2017real} and Mentzer \etal \cite{mentzer2018conditional}. 
Note that since the source codes of the above deep compressors \cite{balle2017end, theis2017lossy, li2017learning, johnston2017improved, rippel2017real, mentzer2018conditional} are not available, we either digitize their rate-distortion curves on the Kodak dataset from the original papers or copy the results from their websites.

\noindent \textbf{Quantitative Evaluation:}
Most of the existing CNN-based LIC models \cite{rippel2017real, mentzer2018conditional, johnston2017improved} are optimized with the MS-SSIM loss \cite{wang2003multiscale}, while traditional LIC methods (\ie, JPEG, JPEG2000 and BPG) and some of the deep LIC methods \cite{balle2017end, theis2017lossy, li2017learning} are optimized in terms of PSNR. 
Therefore, we conduct the experiments to quantitatively evaluate the competing methods in terms of both PSNR and MS-SSIM indices for a more comprehensive comparison.

The PSNR and MS-SSIM based rate-distortion curves on the Kodak and McMaster datasets are summarized in Figure \ref{fig:cur}. As in previous works \cite{toderici2015variable, toderici2017full, balle2017end, theis2017lossy, li2017learning, rippel2017real, mentzer2018conditional, johnston2017improved, agustsson2017soft}, the curves are interpolated based on a set of points [bpp, PNSR] and [bpp, MS-SSIM] for one method. 
Note that for some methods, only the points of [bpp, PNSR] or [bpp, MS-SSIM] are available on the Kodak dataset, and all existing deep LIC methods do not report their results on the McMaster dataset. Therefore, not all methods have all the four curves in Figure \ref{fig:cur}.

Figures \ref{fig:cur}(a) and \ref{fig:cur}(c) show the PSNR based rate-distortion curves on the Kodak and McMaster datasets, respectively. 
One can see that on the Kodak dataset, the proposed TDNet (trained with MSE loss) achieves better result than JPEG2000 and the recently developed deep LIC methods, including Ball{\'e} \etal \cite{balle2017end}, Theis \etal \cite{theis2017lossy} and Li \etal \cite{li2017learning}, and significantly outperforms the prevalent compressor JPEG. 
Although the proposed TDNet does not show advantage over BPG in term of PSNR on the Kodak dataset, it achieves much better PSNR index than BPG on the McMaster dataset (see Figure \ref{fig:cur}(c)). 
Meanwhile, it is not a surprise that TDNet trained with MSE has much higher PSNR indices than TDNet trained with MS-SSIM.

Figures \ref{fig:cur}(b) and \ref{fig:cur}(d) show the MS-SSIM based rate-distortion curves on the Kodak and McMaster datasets, respectively. 
One can see that our TDNet largely outperforms the traditional codecs BPG, JPEG2000 and JPEG. 
It also significantly outperforms the methods of Johnston \etal \cite{johnston2017improved}, Ball{\'e} \etal \cite{balle2017end} and Theis \etal \cite{theis2017lossy}, and achieves comparable performance to the state-of-the-art deep LIC methods Rippel $\&$ Bourdev \cite{rippel2017real} and Mentzer \etal \cite{mentzer2018conditional}.

Again, we would like to stress that all the competing CNN-based LIC methods here train a specific network for a certain bpp, while our proposed DTNet trains a single network to deal with multiple bpp rates.

\vspace{1.2mm}
\noindent \textbf{Visual Quality Evaluation:}
We further compare the visual quality of images compressed by JPEG, JPEG 2000, BPG, Ball{\'e} \etal \cite{balle2017end}, Li \etal \cite{li2017learning}, Theis \etal \cite{theis2017lossy} and our proposed TDNet (trained with MSE and MS-SSIM). 
Note that since the source codes of all existing deep LIC compressors are not available, we can only download the results of Ball{\'e} \etal \cite{balle2017end}, Li \etal \cite{li2017learning} and Theis \etal \cite{theis2017lossy} from their websites. 
The compressed images of other deep LIC methods are not available and thus cannot be compared.  

Figure \ref{fig:re1} shows the compressed images $lighthouse$ by the comparison methods at a compression rate around 0.3bpp (note that Theis \etal, only provides the image $lighthouse$ at 0.375bpp). 
One can see that noticeable blocky and ringing artifacts are inevitable in the reconstructed images by traditional JPEG and JPEG 2000 compression formats. 
While BPG, Theis \etal \cite{theis2017lossy} and Ball{\'e} \etal can produce much better visual quality, they still blur much the edges and over-smooth the textures (see the zoom-in areas). 
Li \etal's method can preserve better the sharp edges and detailed textures, but still generate some noticeable artifacts. 
Compared with these methods, the image compressed by our TDNet method is visually more pleasing with sharper edges and much less artifacts.

Figure \ref{fig:re2} presents the visual comparison results on image $peppers$ from the McMaster dataset at a compression rate around $0.3$bpp. 
Note that since the results on this dataset are not available for all existing deep LIC methods, we only compare TDNet with JPEG, JPEG2000 and BPG. 
Again, one can see noticeable artifacts in the zoom-in areas for the traditional LIC methods. In contrast, the result produced by our proposed TDNet exhibits visually much more pleasing results.

\section{Conclusion And Future Work}
\label{sec:col}
In this paper, we presented a simple yet effective Tucker Decomposition Network (TDNet) with a novel tucker decomposition layer (TDL), which can decompose a latent image representation into a set of matrices and one small core tensor for lossy image compression (LIC). 
By changing the rank of core tensor and its quantization levels, we could easily adjust the bits-per-pixel (bpp) rate of latent image representation, and consequently achieved the goal of using a single CNN model to cover a range of bpp rates. 
An iterative non-uniform quantization scheme was presented to optimize the quantizer, and an all-in-one training strategy was employed to train the TDNet. 
Compared with traditional LIC schemes and previous deep LIC compressors which use different networks to compress images at different bpp rates, our TDNet exhibits very competitive results on benchmark datasets by using a single network.


%

\section*{Acknowledgment}
We gratefully acknowledge the support from NVIDIA Corporation for providing us the Titan X GPU used in this research. 


\ifCLASSOPTIONcaptionsoff
  \newpage
\fi



%
{
\bibliographystyle{IEEEtran}
\bibliography{egbib}
}




\end{document}